\documentclass[journal,twocolumn]{IEEEtran}
\usepackage{url}
\usepackage{algorithm}
\usepackage{algorithmicx}
\usepackage{graphicx}          % Include this line if your 
\graphicspath{{Figures/}}
\usepackage{mathrsfs}         
\usepackage{cite} 
\usepackage{amsmath}   
\usepackage{amsfonts}    
\usepackage{amssymb}   
\usepackage{booktabs}
\usepackage{threeparttable}
\usepackage{graphicx}
\usepackage{comment}
\usepackage{xcolor}
\usepackage[normalem]{ulem}
\newcommand\redsout{\bgroup\markoverwith{\textcolor{red}{\rule[0.5ex]{4pt}{2pt}}}\ULon}
\newtheorem{theorem}{\textbf{Theorem}}

\newtheorem{lemma}{\textbf{Lemma}}

\newtheorem{remark}{\textbf{Remark}}
\newtheorem{example}{\textbf{Example}}
\newtheorem{assumption}{\textbf{Assumption}}

\newcommand{\GJ}[1]{{\color{blue}#1}}

\newcommand*{\QEDA}{\hfill\ensuremath{\blacksquare}}

\IEEEoverridecommandlockouts                              

\title{\LARGE \bf
	Distributed Multi-Agent Reinforcement Learning Based on Graph-Induced Local Value Functions}

\author{Gangshan Jing,~He Bai,~Jemin George,~Aranya Chakrabortty~and~Piyush K. Sharma
	\thanks{G.~Jing is with School of Automation, Chongqing University, Chongqing 400044, China,
		{\tt\small jinggangshan@cqu.edu.cn}}%
	\thanks{H.~Bai is with Oklahoma State University, Stillwater, OK 74078, USA.
		{\tt\small he.bai@okstate.edu}}%
	\thanks{J.~George and P.~Sharma are with the U.S. Army Research Laboratory, Adelphi, MD 20783, USA.
		{\tt\small \{jemin.george,piyush.k.sharma\}.civ@army.mil}}%
	\thanks{A. Chakarabortty is with  North Carolina State University, Raleigh, NC 27695, USA.
		{\tt\small achakra2@ncsu.edu}}
}

\begin{document}
	
	\setlength{\abovedisplayskip}{10pt}
	\setlength{\belowdisplayskip}{10pt}
	
	\maketitle

	\begin{abstract} 
		Achieving distributed reinforcement learning (RL) for large-scale cooperative multi-agent systems (MASs) is challenging because: (i) each agent  has access to only limited information; (ii) issues on scalability and sample efficiency emerge due to the curse of dimensionality. In this paper, we propose a general distributed framework for sample efficient cooperative multi-agent reinforcement learning (MARL) by utilizing the structures of graphs involved in this problem. We introduce three coupling graphs describing three types of inter-agent couplings in MARL, namely, the state graph, observation graph and reward graph. By further considering a communication graph, we propose two distributed RL approaches based on local value functions derived from the coupling graphs. The first approach is able to reduce sample complexity significantly under specific conditions on the aforementioned four graphs. The second approach provides an approximate solution and can be efficient even for problems with dense coupling graphs. Here there is a trade-off between minimizing the approximation error and reducing the computational complexity. Simulations show that our RL algorithms have a significantly improved scalability to large-scale MASs compared with centralized and consensus-based distributed RL algorithms.
	\end{abstract}
	
	%	\vspace{-5mm}
	
	\section{Introduction}
	Reinforcement Learning (RL) \cite{sutton2018reinforcement} aims to find an optimal policy for an agent to accomplish a specific task by making this agent interact with the environment. Although RL has found wide practical applications such as boarding games \cite{silver2017mastering}, robotics \cite{polydoros2017survey}, and power systems \cite{mukherjee2021scalable}, the problem is much more complex for multi-agent reinforcement learning (MARL) due to the non-stationary environment for each agent and the curse of dimensionality. MARL, therefore, has attracted increasing attention recently, and has been studied extensively, see the survey papers \cite{oroojlooyjadid2019review,gronauer2021multi,zhang2021multi}. Nonetheless, many challenges still remain to be overcome. In this paper, we focus on dealing with the following two difficulties in developing distributed cooperative RL algorithms for large-scale networked multi-agent systems (MASs): (i) how to deal with inter-agent couplings all over the network with only local information observable for each agent; and (ii) how to guarantee scalability of the designed RL algorithm to large-scale network systems?
	
	The first challenge refers to the fact that an agent cannot make its decision independently since it will affect and be affected by other agents in different ways. There are mainly three types of structure constraints causing inter-agent couplings in MARL\footnote{In the literature, there have been many different settings for the MARL problem. Our work aims to find a distributed control policy for a dynamic MAS to cooperatively maximize an accumulated global joint reward function. In our literature review, a reference is considered as an MARL reference as long as it employed RL to seek a policy for a group of agents to cooperatively optimize an accumulated global joint function in a dynamic environment.} that have been considered in the literature, e.g., coupled dynamics (transition probability) \cite{qu2020scalable,lin2021multi}, partial observability\footnote{Partial observability generally means that only incomplete information of the environment is observed by the learner. In many numerical experiments for MARL, e.g., \cite{omidshafiei2017deep,nayak2023scalable}, partial observation refers to the observation on partial agents and the environment. In this work, we consider a specific scenario where each agent only observes partial agents, which is consistent with \cite{zhang2020cooperative}.} \cite{omidshafiei2017deep,nayak2023scalable,zhang2020cooperative}, and coupled reward functions \cite{zhang2018fully,lin2021multi,guestrin2002coordinated,kok2006collaborative}. The aforementioned references either consider only one type of structure constraints, or employ only one graph to characterize different types of structure constraints. To give a specific example, consider the MAS as a network of linear systems with dynamics coupling, and set the global objective function as an accumulated quadratic function of state and control policy, the problem of learning an optimal distributed policy becomes a data-driven distributed linear quadratic regulator (LQR) design problem\cite{jing2021asynchronous,li2021distributed}, which involves all the aforementioned structure constraints. In the literature, distributed RL algorithms e.g., \cite{gorges2019distributed,li2021distributed,Jingtcns} have been proposed to deal with this problem. However, the three types of structure constraints are yet to be efficiently utilized. Moreover, the most popular formulation of MARL has long been known as the Markov games \cite{littman1994markov}, while the multi-agent LQR problem \cite{jing2021asynchronous} is only a specific application scenario. 
	
	The scalability issue, as the second challenge, is resulted from high dimensions of state and action spaces of MASs. Although each agent may represent an independent individual, the inter-agent couplings in the MARL problem make it difficult for each agent to learn its optimal policy w.r.t. the global objective with only local information.  In the literature of distributed RL \cite{kar2013cal,zhang2018fully,macua2018diff,zhang2020cooperative}, when dealing with these couplings, the most common method is to make different agents exchange information with each other via a consensus algorithm, so that each agent is able to estimate the value of the global reward function although it only has access to local information from neighbors. However, the performance of such distributed RL algorithms will be similar to or even worse than the centralized RL algorithm because (i) it may take a long time for convergence of consensus when the network is of large scale, (ii) essentially the learning process is conducted via an estimated global reward information. As a result, consensus-based distributed RL algorithms still suffer from significant scalability issues in terms of the convergence rate and the learning variance.

	In this paper, we develop distributed scalable algorithms for a class of cooperative MARL problems where each agent has its individual state space and action space (similar to \cite{qu2020scalable,lin2021multi}), and all the aforementioned three types of couplings exist. We consider a general case where the inter-agent state transition couplings, state observation couplings, and reward couplings are characterized by three different graphs, namely, {\it state graph}, {\it observation graph}, and {\it reward graph}, respectively. Based on these graphs, we derive a {\it learning graph}, which describes the required information flow during the RL process. The learning graph also provides a guidance for constructing a local value function (LVF) for each agent, which is able to play the role of the global value function (GVF)\footnote{Please note that the abbreviation of ``GVF" has been used to denote ``General Value Function" in the literature, e.g., \cite{sutton2011horde}. In this paper, ``GVF" always means global value function.} in learning, but only involves partial agents, and therefore, can enhance scalability of the learning algorithm. 
	
	When each agent has access to all the information involved in LVF, distributed RL can be achieved by policy gradient algorithms immediately. However, this approach is usually based on interactions between many agents, which requires a dense communication graph. To further reduce the number of communication links, we design a distributed RL algorithm based on local consensus, whose computational complexity depends on the aforementioned three graphs (see Theorem \ref{th Lp}). Compared with global consensus-based RL algorithms, local consensus algorithms usually have an improved convergence rate as the network scale is reduced\footnote{Although the convergence rate of a consensus algorithm depends on not only the network scale but also the communication weights, typically the convergence rate can be significantly improved if the network scale is largely reduced.  In \cite{olshevsky2014linear}, the relationship between consensus convergence time and the number of nodes in the network is analyzed under a specific setting for the communication weights.}. This implies that the scalability of this RL algorithm requires specific conditions on the graphs embedded in the MARL problem. To relax the graphical conditions, we further introduce a truncation index and design a truncated LVF (TLVF), which involves fewer agents than LVF. While being applicable to MARL with any graphs, the distributed RL algorithm based on TLVFs only generates an approximate solution, and the approximation error depends on the truncation index to be artificially designed (see Theorem \ref{th Lp kappa}). We will show that there is a trade-off between minimizing the approximation error and reducing the computational complexity (enhancing the convergence rate). 
	
	In \cite{Jing22ACC}, we have considered the case when no couplings exist between the rewards of different individual agents. In contrast, this paper considers coupled individual rewards, which further induces a reward graph. Moreover, this paper provides more interesting graphical results, a distributed RL algorithm via local consensus, and a TLVF-based distributed RL framework.
	
	The main novel contributions of our work that add to the existing literature are summarized as follows.

	(i). We consider a general formulation for distributed RL of networked dynamic agents, where the aforementioned three types of inter-agent couplings exist in the problem, simultaneously. Similar settings have been considered in \cite{qu2020scalable,lin2021multi}. The main novelty here is that the three coupling graphs in this paper are fully independent and inherently exist in the problem. Based on the three graphs corresponding to three types of couplings, we derive a learning graph describing the information flow required in learning. By discussing the relationship between the learning graph and the three coupling graphs (see Lemma \ref{le GL and GCO}), one can clearly observe how different types of couplings affect the required information exchange in learning.
	
	(ii). By employing the learning graph, we construct a LVF for each agent such that the partial gradient of the LVF w.r.t. each individual's policy parameter is exactly the same as that of the GVF (see Lemma \ref{le lg=gg}), which can be directly employed in policy gradient algorithms. MARL algorithms based on LVFs have also been proposed by network-based LVF approximation \cite{qu2020scalable,qu2020scalablenips,lin2021multi} and value function decomposition \cite{guestrin2002coordinated,kok2006collaborative,sunehag2017value,zhang2021fop}. However, the network-based LVF approximation only provided an approximate solution. Moreover, the aforementioned value function decomposition references always assumed that all the agents share a common environment state, therefore never involved the state graph (describing dynamics couplings between different agents).

	(iii). To show the benefits of employing the constructed LVFs in policy gradient algorithms, we focus on zeroth-order optimization (ZOO)\footnote{The ZOO method can be implemented with very limited information (only the objective function evaluations), therefore has a wide range of applications. In recent years, the ZOO-based RL algorithms have been shown efficient in solving model-free optimal control problems, e.g., \cite{Fazel18,malik2020derivative,li2021distributed}. Inspired by these facts, we employ the ZOO-based method to deal with the model-free optimal distributed control problem of MASs under a very general formulation.}. Due to the removal of redundant information (less agents are involved) in gradient estimation, our learning framework based on LVFs always exhibits a reduced gradient estimation variance (see Remark \ref{re variance}) compared with GVF-based policy evaluation. Note that most of the existing distributed ZOO algorithms \cite{Hajine19,Gratton20,Tang20,Akhavan21} essentially evaluate policies via the global value.  %In \cite{tang2020zeroth,Jing21}, LVFs were shown to have reduced variance over GVFs, but how to obtain such LVFs was not discussed.

	(iv). To deal with the scenario when the learning graph is dense, we construct TLVFs by further neglecting the couplings between agents that are far away in the coupling graph. The underlying idea is motivated by \cite{qu2020scalable,qu2020scalablenips,lin2021multi}. Our design, however, is different from them as they construct a TLVF for each agent, whereas we deign a TLVF for each cluster.
	
	The rest of this paper is organized as follows. Section \ref{sec MARL} describes the MARL formulation and the main goal of this paper. Section \ref{sec LVF LG} introduces the LVF design and the learning graph derivation. Section \ref{sec RL LVF} shows the distributed RL algorithm based on LVFs and local consensus, and provides convergence analysis. Section \ref{sec RL TLVF} introduces the RL algorithm based on TLVFs as well as the convergence analysis. Section \ref{sec sim} shows several simulation examples to illustrate the advantages of our proposed algorithms. Section \ref{sec conclusion} concludes the paper. Sections \ref{sec: appendix A} and \ref{sec: appendix B} provide theoretical proofs and the relationships among different cluster-wise graphs, respectively.
	
	\textbf{Notation}: Throughout the paper, unless otherwise stated, $\mathcal{G}_X=(\mathcal{V},\mathcal{E}_X)$ always denotes an unweighted directed graph\footnote{In this paper, we will introduce multiple graphs, here $X$ may represent $S$, $O$, $C$, $R$ and $L$.}, where $\mathcal{V}=\{1,...,N\}$ is the set of vertices, $\mathcal{E}_X\subset\mathcal{V}\times\mathcal{V}$ is the set of edges, $(i,j)\in\mathcal{E}_X$ means that there is a directional edge in $\mathcal{G}$ from $i$ to $j$. The in-neighbor set and out-neighbor set of agent $i$ are denoted by $\mathcal{N}_i^X=\{j\in\mathcal{V}:(j,i)\in\mathcal{E}_X\}$, and $\mathcal{N}_i^{X+}=\{j\in\mathcal{V}:(i,j)\in\mathcal{E}_X\}$, respectively. A path from $i$ to $j$ is a sequence of distinct edges of the form $(i_1,i_2)$, $(i_2,i_3)$, ..., $(i_{r-1},i_r)$ where $i_1=i$ and $i_r=j$. We use $i\stackrel{\mathcal{E}}{\longrightarrow} j$ to denote that there is a path from $i$ to $j$ in edge set $\mathcal{E}$. A subgraph $\mathcal{G}'=(\mathcal{V}',\mathcal{E}')$ with $\mathcal{V}'\subseteq\mathcal{V}$ and $\mathcal{E}'\subseteq\mathcal{E}$ is said to be a strongly connected component (SCC) if there is a path between any two vertices in $\mathcal{G}'$. One vertex is a special SCC. Given a directed graph $\mathcal{G}=(\mathcal{V},\mathcal{E})$, we define the {\it transpose graph} of $\mathcal{G}$ as $\mathcal{G}^\top=(\mathcal{V},\mathcal{E}^\top)$, where $\mathcal{E}^\top=\{(i,j)\in\mathcal{V}\times\mathcal{V}: (j,i)\in\mathcal{E}\}$. Given two edge sets $\mathcal{E}_1$ and $\mathcal{E}_2$, it can be verified that $\mathcal{E}_1\subseteq\mathcal{E}_2$ if and only if $\mathcal{E}_1^\top\subseteq\mathcal{E}_2^\top$. Moreover, if $(i,j), (j,i)\in\mathcal{E}$, then $(i,j),(j,i)\in\mathcal{E}^\top$. Given a set $A$ and a vector $v$, $v_A=(...,v_i,...)^\top$ with $i\in A$. Given a set $X$, $\mathcal{P}(X)$ is the set of probability distributions over $X$. The $d\times d$ identity matrix is denoted by $I_d$, the $a\times b$ zero matrix is denoted by $\mathbf{0}_{a\times b}$. $\mathbb{R}^d$ is the $d$-dimensional Euclidean space. $\mathbb{N}$ is the set of non-negative integers.

	\section{Multi-Agent Reinforcement Learning}\label{sec MARL}
	
	Consider the optimal control problem of a MAS modeled by a Markov decision process (MDP), which is described as a tuple $\mathcal{M}=(\mathcal{G}_{\{S,O,R,C\}}, \Pi_{i\in\mathcal{V}}\mathcal{T}_i, \Pi_{i\in\mathcal{V}}\mathcal{O}_i,\Pi_{i\in\mathcal{V}}r_i,\gamma)$ with $\mathcal{G}_{\{S,O,R,C\}}=(\mathcal{V},\mathcal{E}_{\{S,O,R,C\}})$ describing different interaction graphs and $\mathcal{T}_i=(\mathcal{S}_i, \mathcal{A}_i,\mathcal{P}_i)$ specifying the evolution process\footnote{The evolution process of each agent depends on other agents, therefore does not possess the Markov property. However, the whole MAS has the Markov property as its full state only depends on the state and action at last step, and is independent of previous states and actions.} of agent $i$. Detailed notation explanations are listed below:
	\begin{itemize}
		\item $\mathcal{V}=\{1,...,N\}$ is the set of agent indices;
		
		\item $\mathcal{E}_S\subseteq\mathcal{V}\times\mathcal{V}$ is the edge set of the {\it state graph} $\mathcal{G}_S=(\mathcal{V},\mathcal{E}_S)$, which specifies the dynamics couplings among the agents' states, $(i,j)\in\mathcal{E}_S$ implies that the state evolution of agent $j$ involves the state of agent $i$;
		
		\item $\mathcal{E}_O\subseteq\mathcal{V}\times\mathcal{V}$ is the edge set of the {\it observation graph} $\mathcal{G}_O=(\mathcal{V},\mathcal{E}_O)$, which determines the partial observation of each agent. More specifically, agent $i$ observes the state of agent $j$ if $(j,i)\in\mathcal{E}_O$;
		
		\item $\mathcal{E}_R\subseteq\mathcal{V}\times\mathcal{V}$ is the edge set of the {\it reward graph} $\mathcal{G}_R=(\mathcal{V},\mathcal{E}_R)$, which describes inter-agent couplings in the reward of each individual agent, the reward of agent $i$ involves the state and the action of agent $j$ if $(j,i)\in\mathcal{E}_R$;
		
		\item $\mathcal{E}_C\subseteq\mathcal{V}\times\mathcal{V}$ is the edge set of the {\it communication graph} $\mathcal{G}_C=(\mathcal{V},\mathcal{E}_C)$. An edge $(i,j)\in\mathcal{E}_C$ implies that agent $j$ is able to receive information from agent $i$;
		
		\item $\mathcal{S}_i$ and $\mathcal{A}_i$ are the state space and the action space of agent $i$, respectively, and can be either continuous or finite;
		
		\item $\mathcal{P}_i:\Pi_{j\in\mathcal{I}_i^S}\mathcal{S}_j\times \Pi_{j\in\mathcal{I}_i^S}\mathcal{A}_j\rightarrow \mathcal{P}(\mathcal{S}_i)$ is the transition probability function specifying the state probability distribution of agent $i$ at the next time step under current states $\{s_j\}_{j\in\mathcal{I}_i^S}$ and actions $\{a_j\}_{j\in\mathcal{I}_i^S}$, here $\mathcal{I}_i^S=\{j\in\mathcal{V}:(j,i)\in\mathcal{E}_S\}\cup\{i\}$ includes agent $i$ and its neighbors in the state graph $\mathcal{G}_S$;
		
		\item $r_i:\Pi_{j\in\mathcal{I}_i^R}\mathcal{S}_j\times\Pi_{j\in\mathcal{I}_i^R}\mathcal{A}_j\rightarrow\mathbb{R}$ is the immediate reward returned to agent $i$ when each agent $j\in\mathcal{I}^R_i$ takes action $a_j\in\mathcal{A}_j$ at the current state $s_j\in\mathcal{S}_j$, here $\mathcal{I}_i^R=\{j\in\mathcal{V}:(j,i)\in\mathcal{E}_R\}\cup\{i\}$;
		
		\item $\mathcal{O}_i=\{\Pi_{j\in\mathcal{I}_i^O}\mathcal{S}_j\}$ is the observation space of agent $i$, which includes the states of all the agents in $\mathcal{I}_i^O$, here $\mathcal{I}_i^O=\{j\in\mathcal{V}:(j,i)\in\mathcal{E}_O\}\cup\{i\}$;
		
		\item $\gamma\in(0,1)$ is the discount factor that trades off the instantaneous and future rewards.
	\end{itemize}

	Let $\mathcal{S}=\Pi_{j\in\mathcal{V}}\mathcal{S}_j$, $\mathcal{A}=\Pi_{j\in\mathcal{V}}\mathcal{A}_j$, and $\mathcal{P}=\Pi_{j\in\mathcal{V}}\mathcal{P}_j$ denote the joint state space, action space and transition probability function of the whole MAS. Each agent $i$ has a state $s_i\in\mathcal{S}_i$ and an action $a_i\in\mathcal{A}_i$. The global state and action at time step $t$ are denoted by $s(t)=(s_1(t),...,s_N(t))$ and $a(t)=(a_1(t),...,a_N(t))$, respectively.
	Let $\pi: \mathcal{S}\rightarrow \mathcal{P}(\mathcal{A})$ and $\pi_i: \mathcal{O}_i\rightarrow \mathcal{P}(\mathcal{A}_i)$ be a global policy function of the MAS and a local policy function of agent $i$, respectively. Here $\mathcal{P}(\mathcal{A})$ and $\mathcal{P}(\mathcal{A}_i)$ are the sets of probability distributions over $\mathcal{A}$ and $\mathcal{A}_i$, respectively. The global policy is the policy of the whole MAS from the centralized perspective, thus is based on the global state $s$. The local policy of agent $i$ is based on the local observation $o_i=s_{\mathcal{I}_i^O}\in\mathcal{O}_i$, which constitutes states of partial agents. Note that a global policy always corresponds to a collection of local policies uniquely.
	
	At each time step $t$ of the MDP, each agent $i\in\mathcal{V}$ executes an action $a_{i}(t)\in\mathcal{A}_i$ according to its policy $\pi_i$ and the local observation $o_{i}(t)=s_{\mathcal{I}_i^O}(t)\in\mathcal{O}_i$, then obtains a reward $r_i(s_{\mathcal{I}_i^R}(t),a_{\mathcal{I}_i^R}(t))$. Note that such a formulation is different from that in \cite{zhang2018fully,chen2021communication}, where the transition and reward of each agent are associated with the global state $s(t)$. Moreover, different from many MARL references where the reward of each agent only depends on its own state and action, in our work, we consider a more general formulation for cooperative MARL where the reward of each agent may be influenced by other agents, determined by the reward graph $\mathcal{G}_R=(\mathcal{V},\mathcal{E}_R)$.

	%From the definition, each agent has its independent state space, action space, and transition probability. Given local policy $\pi_i$, each agent takes its action based on the local observation $o_i$, which is independent of  its neighbors' states, in contrast to the global observation assumption in \cite{zhang2018fully,chen2021communication}.  Moreover, different agents are coupled in their local rewards, meaning that the reward returned to each agent depends on not only its own action, but also its collaborative neighbors' actions. Intuitively, the assumption on local reward here is more restrictive than the independent local reward assumption in \cite{zhang2018fully}, it is, however, valid in many application scenarios. For example, a group of robots aim to cooperatively move supplies from other locations to location $A$. The amount of supplies at location $A$ can be the common reward for all the robots, and it certainly depends on all the robots' actions.

	The long-term accumulated discounted global reward is defined as
	\begin{equation}R=\sum_{t=0}^{\infty}\gamma^tr(s(t),a(t))=\sum_{i=1}^N\sum_{t=0}^{\infty}\gamma^tr_{i}(s_{\mathcal{I}_i^R}(t),a_{\mathcal{I}_i^R}(t)),
	\end{equation}
where $r(s(t),a(t))$ is the global reward for the MAS at time $t$, $r_i$ is the local reward for agent $i$ at  time $t$. Note that maximizing $R$ is equivalent to maximizing its average $\frac{1}{N}R$, which has been commonly adopted as the learning objective in many MARL references, e.g. \cite{kar2013cal,zhang2018fully,zhang2020cooperative}. Based on this long term global reward, with a given policy $\pi$, we are able to define the global state value function $V^\pi(s)=\mathbb{E}[\sum_{t=0}^{\infty}\gamma^tr(s(t),a(t))|s(0)=s]$ and state-action value function $Q^\pi(s,a)=\mathbb{E}[\sum_{t=0}^{\infty}\gamma^tr(s(t),a(t))|s(0)=s,a(0)=a]$, which describe the expected long term global reward when agents have initial state $s$ and initial state-action pair $(s,a)$, respectively. Similarly, the local state value function with initial state $s$ for each agent $i$ can be defined as $V_i^\pi(s)=\mathbb{E}[\sum_{t=0}^{\infty}\gamma^tr_i(s_{\mathcal{I}_i^R}(t),a_{\mathcal{I}_i^R}(t))|s(0)=s]$.

	The goal of this paper is to design a distributed RL algorithm for the MAS to find a control policy $\pi$ maximizing $J(\pi)=\mathbb{E}_{s\sim\mathcal{D}}V^{\pi}(s)$, whose expression is
	\begin{equation}\label{RL objective}
			\begin{split}
				&\mathbb{E}_{s\sim\mathcal{D}}\left[\sum_{i=1}^N\sum_{t=0}^\infty \mathbb{E}_{a(t)\sim\pi(s(t))}\gamma^tr_i(s_{\mathcal{I}_i^R}(t),a_{\mathcal{I}_i^R}(t))|s(0)=s\right],
			\end{split}
	\end{equation}
where $\mathcal{D}$ denotes the distribution that the initial state follows. For convenience of analysis, we also define the expected value to be maximized corresponding to individual reward for each agent $i$ as 
	\begin{equation}
			J_i(\pi)=\mathbb{E}_{s\sim\mathcal{D}}V_i^\pi(s), i\in\mathcal{V}.
	\end{equation}
Note that here $J_i$ may be determined by the policies of partial agents, instead of the global policy $\pi$. However, the global policy $\pi$ is always able to determine $J_i$, therefore, can be employed as the argument of $J_i$.

	We parameterize the global policy $\pi(s,a)$ using parameters $\theta=(\theta_1^\top,...,\theta_N^\top)^\top\in\mathbb{R}^d$ with $\theta_i\in\mathbb{R}^{d_i}$. The global policy and agent $i$'s local policy are then rewritten as $\pi^\theta(s,a)$ and $\pi^{\theta_i}_i(o_i,a_i)$, respectively. Note that given any global state $s\in\mathcal{S}$, a global policy and a collection of local policies can always be transformed to each other. Now we turn to solve the following optimization problem:
	\begin{equation}\label{maxJ}
			\max_{\theta} J(\theta):=\mathbb{E}_{s\sim\mathcal{D}}V^{\pi(\theta)}(s).
	\end{equation}
	%which is an approximation of (\ref{RL objective}) because the feasible domain of $\pi$ is restricted to parameterized functions.

	Next we present a distributed multi-warehouse resource transfer problem to demonstrate our formulation. This example is a variation of many practical applications, e.g., it can also be read as a problem of energy transfer among different rooms in a smart building.
	
\begin{figure}%[htbp]
		\centering
		\includegraphics[width=9cm]{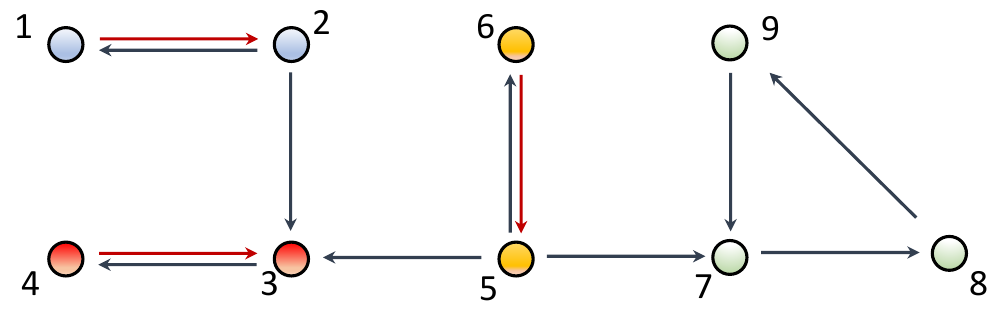}
		\caption{The state graph $\mathcal{G}_S=(\mathcal{V},\mathcal{E}_S)$ and the observation graph $\mathcal{G}_O=(\mathcal{V},\mathcal{E}_O)$. The black and red lines correspond to edges in $\mathcal{E}_S$ and $\mathcal{E}_O$, respectively.} \label{fig GCO}
	\end{figure}

\begin{figure}%[htbp]
		\centering
		\includegraphics[width=9cm]{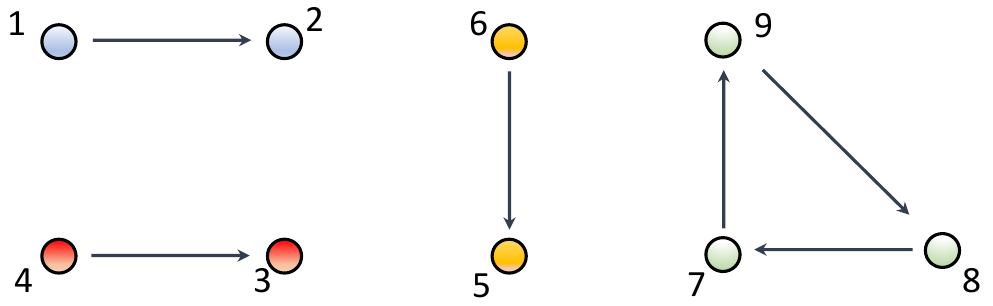}
		\caption{The reward graph $\mathcal{G}_R=(\mathcal{V},\mathcal{E}_R)$.} \label{fig GR}
	\end{figure}

	\begin{example}\label{ex warehouse}
		Consider a network of 9 warehouses $\mathcal{V}=\{1,...,9\}$ consuming resources while transferring resources among each other. The goal is to guarantee adequate supplies for each warehouse. Each warehouse is denoted by a vertex in the graph. The state graph $\mathcal{G}_S=(\mathcal{V},\mathcal{E}_S)$ interpreting the transition relationship is shown in Fig. \ref{fig GCO}. The observation graph $\mathcal{G}_O$ only contains 3 edges involving 3 leaf nodes in graph $\mathcal{G}_S$, as shown in Fig. \ref{fig GCO}, which implies that only warehouses 2, 3, and 5 observe the current resource stock of warehouses other than itself. The motivation behind this setting is that warehouses 1, 4 and 6 do not send out resources at all, hence their neighbors need to keep monitoring their states so that the resources sent to them are neither insufficient nor redundant. The reward graph $\mathcal{G}_R$ is shown in Fig. \ref{fig GR}, which contains $\mathcal{G}_O$ as a subgraph. This ensures that the observation of each warehouse always influences its own reward, implying that a warehouse is responsible for the resource shortage of those warehouses it can observe. At time step $t$, warehouse $i\in\mathcal{V}$ stores resources of the amount $m_i(t)\in\mathbb{R}$,  receives a local demand $d_i(t)\in\mathbb{R}$, sends partial of its resources to and receives resources from its neighbors $j\in\mathcal{N}_i^S$ in the state graph $\mathcal{G}_S$, besides its neighbors, warehouse $i$ also receives resources supply of the amount $y_i(t)$ from outside. Let $z_i(t)=y_i(t)-d_i(t)$, then agent $i$ has the following dynamics
			\begin{align}\label{dynamics}
				m_i(t+1)&=m_i(t)-\sum_{j\in\mathcal{N}_i^{out}}b_{ij}(o_i(t))\alpha_im_i(t)\nonumber\\
				&~~~~+\sum_{j\in\mathcal{N}_i^C}b_{ji}(o_j(t))\alpha_jm_j(t)+z_i(t),\nonumber\\
				z_i(t)&=A_i\sin(w_{i}t+\phi_i)+\omega_i,
			\end{align}
		where $b_{ij}(o_i(t))\in[0,1]$ denotes the fraction of resources agent $i$ sends to its neighbor $j$ at time $t$, $\alpha_i$ determines whether the $i$-th warehouse has resources to send out, i.e., $\alpha_i=0$ if $m_i\leq0$, and $\alpha_i=1$ otherwise, $0<A_i<m_i(0)$ is a constant, $w_{i}$ is a bounded random quantity, $i\in\mathcal{V}$ and $\phi$ is a positive scalar, $\mathcal{N}_i^S=\{j\in\mathcal{V}: (j,i)\in\mathcal{E}_S\}$ and  $\mathcal{N}_i^{S+}=\{j\in\mathcal{V}:(i,j)\in\mathcal{E}_S\}$ are the in-neighbor set and the out-neighbor set of agent $i$, respectively.
		
		From the MARL perspective, besides the three graphs and transition dynamics introduced above, the rest of entries in $\mathcal{M}$ for each agent $i$ at time step $t$ can be recognized as
		\textbf{Individual state:} $s_i(t)=(m_i(t),z_i(t))^\top$. \textbf{Individual action:} $a_i(t)=(...,b_{ij}(o_i(t)),...)^\top_{j\in\mathcal{N}_i^{S+}}$. \textbf{Individual policy function:} $\pi_i(\cdot)=(...,b_{ij}(\cdot),...)^\top_{j\in\mathcal{N}_i^{S+}}$.
		\textbf{Partial observation:} $o_{i}(t)=(\{m_j(t)\}_{j\in\mathcal{I}_i^O},z_i(t))^\top\in\mathbb{R}^{|\mathcal{I}_i^O|+1}$.
		\textbf{Individual reward:} $r_{i}(t)=\sum_{j\in\mathcal{I}_i^R}\tau_j(t)$, where $\tau_j(t)=0$ if $m_j(t)\geq0$, and $\tau_{j}(t)=-m_j^2(t)$ otherwise.
		
		The goal of the resource transfer problem is to maximize $\mathbb{E}_{s(0)\sim\mathcal{D}}\sum_{i=1}^N\sum_{t=0}^{\infty}\gamma^tr_i(t)$ under the dynamics constraint (\ref{dynamics}). In other words, we aim to find the optimal transfer policy such that each warehouse keeps having enough resources for use.
		
		\begin{remark}
			Note that many settings in this example can be adjusted while maintaining the applicability of the proposed approach in this paper. For example, the partial observation of each agent $i$ can be replaced by $o_i=(...,m_j(t), d_j(t),...)^\top_{j\in\mathcal{I}_i^O}$ or $o_i=(...,m_j(t),...)^\top_{j\in\mathcal{I}_i^O}$. Depending on different observation settings, the optimal policy may change.
		\end{remark}
	\end{example}
	
	Existing distributed policy gradient methods such as actor-critic~\cite{zhang2018fully}  and zeroth-order optimization \cite{zhang2020cooperative} can be employed to solve the problem when there is a connected undirected communication graph among the agents. However, these approaches are based on estimation of the GVF, which requires a large amount of communications during each learning episode. Moreover, policy evaluation based on the GVF has a significant scalability issue due to the high dimension of the state and action spaces for large-scale networks.

	\section{Local Value Function and Learning Graph}\label{sec LVF LG}
	
	In this section, we introduce how to design an appropriate LVF for each agent, which involves only partial agents, but its gradient w.r.t. the local policy parameter is the same as that of the GVF.
	
	\subsection{Local Value Function Design}\label{subsec LVF design}
	
	Although the state graph $\mathcal{G}_S$, the observation graph $\mathcal{G}_O$, and the reward graph $\mathcal{G}_R$ can be defined independently, we observe that all of them will induce the couplings between different agents in the optimization objective. In this subsection, we will build a connection between these graphs and the couplings between agents, based on which the LVFs will be designed.
	
	Define a new graph $\mathcal{G}_{SO}\triangleq(\mathcal{V},\mathcal{E}_{SO})$ where $\mathcal{E}_{SO}=\mathcal{E}_S\cup\mathcal{E}_O$, and define 
	\begin{equation}\label{R_i^SO}
			\mathcal{R}^{SO}_i=\{j\in\mathcal{V}: i\stackrel{\mathcal{E}_{SO}}{\longrightarrow} j\}\cup\{i\},
	\end{equation}
which includes the vertices in graph $\mathcal{G}_{SO}$ that are reachable from vertex $i$ and vertex $i$ itself. In fact, the states of the agents in $\mathcal{R}_i^{SO}$ will be affected by agent $i$'s state and action as time goes on.
	
	To design the LVF for each agent $i$, we need to specify the agents whose individual rewards will be affected by the action of agent $i$. To this end, we define the following composite reward for agent $i$:
	\begin{equation}
			\hat{r}_i(s(t),a(t))=\sum_{j\in\mathcal{I}_i^L}r_j(s_{\mathcal{I}_j^R}(t),a_{\mathcal{I}_j^R}(t)),
	\end{equation}
where 
	\begin{equation}\label{I_i^L}
			\mathcal{I}_i^L=\{j\in\mathcal{V}: \mathcal{I}_j^R\cap\mathcal{R}_i^{SO}\neq\varnothing\}=\cup_{k\in\mathcal{R}_i^{SO}}\mathcal{I}_k^{R+},
	\end{equation}here $\mathcal{I}_k^{R+}=\{j\in\mathcal{V}:(k,j)\in\mathcal{E}_R\}\cup\{k\}$ consists of the out-neighbors of vertex $k$ in graph $\mathcal{G}_R$ and itself.
	
	To demonstrate the definitions of $\mathcal{R}_i^{SO}$ and $\mathcal{I}_i^L$, let us look at Example \ref{ex warehouse}. One can observe from Fig. \ref{fig GCO} and Fig. \ref{fig GR} that $\mathcal{R}_1^{SO}=\mathcal{R}_2^{SO}=\mathcal{I}_1^L=\mathcal{I}_2^L=\{1,2,3,4\}$. In fact, we have $\mathcal{R}_i^{SO}=\mathcal{I}_i^L$ since $\mathcal{I}_k^{R+}\subset \mathcal{R}_i^{SO}$ for all $k\in\mathcal{R}_i^{SO}$, $i\in\mathcal{V}$.
	
	Accordingly, we define the LVF for agent $i$ as
	\begin{equation}\label{local value}
			\hat{V}^\pi_i(s)=\mathbb{E}\left[\sum_{t=0}^\infty\gamma^t\hat{r}_i(s(t),a(t))|s(0)=s\right].
	\end{equation}
	
	When the GVF is replaced by the LVF, the agent $i$ is expected to maximize the following objective:
	\begin{equation}\label{Jhati}\hat{J}_i(\theta)=\mathbb{E}_{s\sim\mathcal{D}}\hat{V}^{\pi(\theta)}_i(s)=\sum_{j\in\mathcal{I}^L_i}J_j(\theta).
	\end{equation}
	Different from the global objective function $J(\theta)=\sum_{j\in\mathcal{V}}J_j(\theta)$, the local objective $\hat{J}_i(\theta)$ only involves agents in a subset $\mathcal{I}_i^L\subseteq\mathcal{V}$. We make the following assumption on the graphs so that $\hat{J}_i(\theta)\neq J(\theta)$ for at least one agent $i$.
	\begin{assumption}\label{as weakly connected}
		There exists a vertex $i\in\mathcal{V}$ such that $\mathcal{I}_i^L\neq\mathcal{V}$.	
	\end{assumption}
	
	Define graph $\mathcal{G}_{SOR}=\mathcal{G}_{SO}\cup\mathcal{G}_R$. The following lemma shows a sufficient graphical condition and a necessary graphical condition for Assumption \ref{as weakly connected}.
	
	\begin{lemma}\label{le GCOR}
		The following statements are true:
		
		(i). Assumption \ref{as weakly connected} holds if graph $\mathcal{G}_{SOR}$ has $n>1$ SCCs.
		
		(ii). Assumption \ref{as weakly connected} holds only if graph $\mathcal{G}_{SO}$ has $n>1$ SCCs.
	\end{lemma}
	
\begin{figure}%[htbp]
		\centering
		\includegraphics[width=9cm]{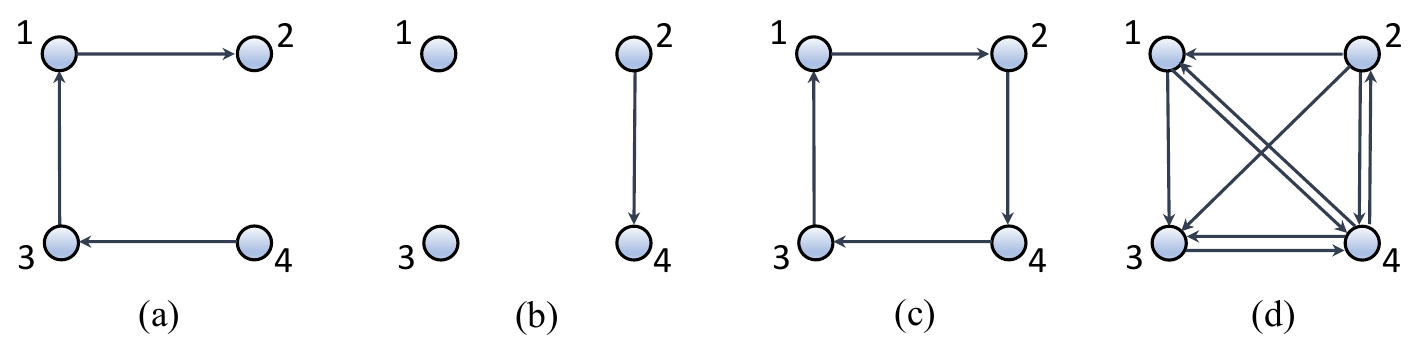}
		\caption{A counter-example for the converse of Lemma \ref{le GCOR} (i). Graphs (a), (b), (c), and (d) denote graphs $\mathcal{G}_{SO}$, $\mathcal{G}_R$, $\mathcal{G}_{SOR}$, and the learning graph $\mathcal{G}_L$, respectively.} \label{fig counterexample}
	\end{figure}

	One may question if the converses of the statements in Lemma \ref{le GCOR} are true. Both answers are no. This is because graph $\mathcal{G}_R$ may contain some edges that connect different SCCs in $\mathcal{G}_{SO}$, but the paths involving more than two vertices in $\mathcal{G}_R$ cannot be used in expanding $\mathcal{I}_i^R$. For statement (i), $\mathcal{G}_{SOR}$ may be strongly connected even when there exists a vertex $j\in\mathcal{V}\setminus\mathcal{I}_i^L$. Fig. \ref{fig counterexample} shows a counter-example where $\mathcal{I}_1^L=\{1,2,4\}$ is only a subset of $\mathcal{V}$ but $\mathcal{G}_{SOR}$ is strongly connected. For statement (ii), a simple counter-example can be obtained by setting  $\mathcal{E}_R=\mathcal{V}\times\mathcal{V}$. Note that Lemma 1 induces a necessary and sufficient condition for Assumption \ref{as weakly connected} when $\mathcal{G}_{SO}=\mathcal{G}_{SOR}$, which happens when $\mathcal{G}_R\subseteq \mathcal{G}_{SO}$.

	%\vspace{-0.1cm}

	Next we show that each agent $i\in\mathcal{V}$ maximizing (\ref{Jhati}) is equivalent to maximizing the global objective (\ref{maxJ}).

	Given a function $f(\theta):\mathbb{R}^d\rightarrow\mathbb{R}$ and a positive $\delta$, we define
	\begin{equation}
			f^\delta(\theta)=\mathbb{E}[f(\theta+\delta u)], ~~u\sim\mathcal{N}(0,I_d).
	\end{equation}
	The following lemma shows the equivalence between the gradients of the smoothed local objective  $\hat{J}_i^\delta(\theta)$ and the smoothed global objective $J^\delta(\theta)$ w.r.t. the local policy parameter of each individual agent.

	\begin{lemma}\label{le lg=gg}
		The following statements are true: 
		
		(i)	$\nabla_{\theta_i}J^\delta(\theta)=\nabla_{\theta_i}\hat{J}_i^\delta(\theta)$ for any $\delta>0$, $i\in\mathcal{V}$.
		
		(ii) If $J_i(\theta)$, $i\in\mathcal{V}$ are differentiable, then $\nabla_{\theta_i}J(\theta)=\nabla_{\theta_i}\hat{J}_i(\theta)$, $i\in\mathcal{V}$.
	\end{lemma}

	Lemma \ref{le lg=gg} reveals the implicit connection between graphs and agents' couplings in the optimization objective, and provides the theoretical guarantee for the reasonability of the designed LVFs. It is important to give the following two notes. (a). Although the RL algorithm in this paper is based on ZOO, Lemma \ref{le lg=gg} is independent of ZOO. Therefore, Lemma \ref{le lg=gg} is also compatible with other policy gradient algorithms. (b). Statement (i) in Lemma \ref{le lg=gg} does not require $J_i(\theta)$, $i\in\mathcal{V}$ to be differentiable because $J^\delta(\theta)=\mathbb{E}[J(\theta+\delta u)]$ is always differentiable \cite[Lemma 2]{nesterov2017random}. 
	
	In order to adapt our approach to the scenario when $J(\theta)$ is not differentiable, we choose to find the stationary point of $J^\delta(\theta)$. The gap between $J(\theta)$ and $J^\delta(\theta)$ can be bounded if $J(\theta)$ is Lipschitz continuous and $\delta>0$ is sufficiently small.
	
	To guarantee the Lipschitz continuity\footnote{The Lipschitz continuity of a value function implies that similar policy parameters have similar performance for the problem. This is reasonable in practice especially for problems with continuous state and action spaces. In \cite{pirotta2015policy}, it has been shown that the value function becomes Lipschitz continuous w.r.t. policy parameters as long as both the MDP and the policy function have Lipschitz continuity properties.} of $J(\theta)$, we make the following assumption on functions $V_i^{\pi(\theta)}(s)$ for $i\in\mathcal{V}$:
	\begin{assumption}\label{as Lip}
		$V_i^{\pi(\theta)}(s)$, $i\in\mathcal{V}$ are $L_i$-Lipschitz continuous w.r.t. $\theta$ in $\mathbb{R}^d$ for any $s\in\mathcal{S}$. That is, $|V_i^{\pi(\theta)}(s)-V_i^{\pi(\theta')}(s)|\leq L_i\|\theta-\theta'\|$ for any $s\in\mathcal{S}$, $\theta,\theta'\in\mathbb{R}^d$.
	\end{assumption}

	Assumption \ref{as Lip} directly implies that $J_i(\theta)$ is $L_i$-Lipschitz continuous. Moreover, $J(\theta)$ is $L$-Lipschitz continuous in $\mathbb{R}^d$, where $L\triangleq\sum_{i\in\mathcal{V}}L_i$, due to the following fact:
	\begin{equation}
			\begin{split}
				&|J(\theta)-J(\theta')|\leq\sum_{i\in\mathcal{V}}|J_i(\theta)-J_i(\theta')|\\
				&\leq \sum_{i\in\mathcal{V}}\mathbb{E}\left[\left|V_i^{\pi(\theta)}(s)-V_i^{\pi(\theta')}(s)\right|\right]\leq \sum_{i\in\mathcal{V}}L_i\|\theta-\theta'\|.
			\end{split}
	\end{equation}
	
	\subsection{Learning Graph}\label{subsec: learning graph}
	
	Lemma \ref{le lg=gg} has shown that having the local gradient of a specific local objective function is sufficient for each agent to optimize its policy according to the following gradient ascent:
	\begin{equation}\label{gradient ascent}	\theta_i^{k+1}=\theta_i^k+\nabla_{\theta_i}\hat{J}_i^\delta(\theta^k),~~~~i\in\mathcal{V},
	\end{equation}
where $\theta^k$ is the policy parameter at step $k$, and $\nabla_{\theta_i}\hat{J}_i^\delta(\theta^k)$ can be estimated by evaluating the value of $\hat{J}_i^\delta(\theta^k)$.

	Then we are able to define the {\it learning graph} $\mathcal{G}_L=(\mathcal{V},\mathcal{E}_L)$ based on the set $\mathcal{I}_i^L$ (\ref{I_i^L}) in the LVF design, which interprets the required reward information flow during the learning process. The edge set $\mathcal{E}_L$ is defined as:
	\begin{equation}\label{EL}
			\mathcal{E}_L=\{(j,i)\in\mathcal{V}\times\mathcal{V}: j\in\mathcal{I}_i^L,i\in\mathcal{V}\}.
	\end{equation}
	
	The definition of $\mathcal{E}_L$ implies the following result.
	\begin{lemma}\label{le ELtoECOR}
		If $(j,i)\in\mathcal{E}_L$, then $i\stackrel{\mathcal{E}_{SOR}}{\longrightarrow} j$.
	\end{lemma}
	The converse of Lemma \ref{le ELtoECOR} is not true, see Fig. \ref{fig counterexample} as a counterexample. More specifically, $1\stackrel{\mathcal{E}_{SOR}}{\longrightarrow} 3$, as shown in graph (c), however, $(3,1)\notin\mathcal{E}_L$, see the learning graph $(d)$.

	To better understand the learning graph $\mathcal{G}_L$, we find a clustering $\mathcal{V}=\cup_{l=1}^n\mathcal{V}_l$ for the graph $\mathcal{G}_{SO}$, where $\mathcal{V}_l$ is the vertex set of the $l$-th maximal strongly connected component (SCC) in $\mathcal{G}_{SO}$, and $\mathcal{V}_{l_1}\cap\mathcal{V}_{l_2}=\varnothing$ for any distinct $l_1,l_2\in\mathcal{C}=\{1,...,n\}$. According to Lemma \ref{le GCOR}, such a clustering with $n>1$ can always be  found under Assumption \ref{as weakly connected}.
	
	According to the definitions (\ref{I_i^L}) and (\ref{EL}), we have the following observations:
	\begin{itemize}
		\item The agents in each cluster $\mathcal{V}_l$ form a clique in $\mathcal{G}_L$.
		
		%\item If there is one link in $\mathcal{G}_L$ from one cluster to another cluster, then there is an edge in $\mathcal{G}_L$ between any pair of agents in these two clusters.
		
		\item  The agents in the same cluster share the same LVF.
	\end{itemize}
	
	The first observation holds because any pair of agents in each cluster are reachable to each other and $(j,i)\in\mathcal{E}_L$ as long as $j$ is reachable from $i$ in graph $\mathcal{G}_{SO}$. The second observation holds because $\mathcal{R}_i^{SO}=\mathcal{R}_j^{SO}$ for any $i$ and $j$ belonging to the same cluster, here $\mathcal{R}_i^{SO}$ is defined in (\ref{R_i^SO}).
	
\begin{figure}
		\centering
		\includegraphics[width=8cm]{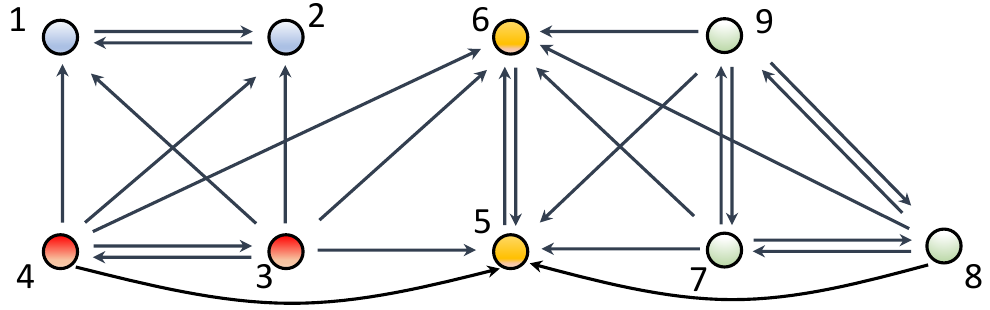}
		\caption{The learning graph $\mathcal{G}_L$ corresponding to $\mathcal{G}_{SO}$ in Fig. \ref{fig GCO} and $\mathcal{G}_R$ in Fig. \ref{fig GR}.} \label{fig learning graph}
	\end{figure}

	To demonstrate the edge set definition (\ref{EL}), the learning graph corresponding to the state graph and the observation graph in Fig. \ref{fig GCO}, and the reward graph in Fig. \ref{fig GR}, is shown in Fig. \ref{fig learning graph}. In fact, it is interesting to see some connections between different graphs from the cluster-wise perspective. Please refer to Appendix B for more details.
	
	%In real applications, one coordinator can be chosen in each cluster, based on which different clusters are able to exchange information with each other, so that a large amount of communication links between different clusters in Fig. \ref{fig learning graph} can be avoided. 
	
	The learning graph $\mathcal{G}_L$ interprets the required reward information flow in our distributed MARL algorithm. If the agents are able to exchange information via communications following $\mathcal{G}_L$, then each agent can acquire the information of its LVF via local communications with others. The zeroth-order oracle in \cite{nesterov2017random} can then be employed to estimate $\nabla_{\theta_i}\hat{J}_i^\delta(\theta^k)$ in (\ref{gradient ascent}). However, $\mathcal{G}_L$ is usually dense, inducing high communication costs, and having such a dense communication graph may be unrealistic in practice. To further relax the condition on the communication graph, in the next section, we will design a distributed RL algorithm based on local consensus algorithms.

	\section{Distributed RL Based on Local Consensus}\label{sec RL LVF}
	
	In this section, we propose a distributed RL algorithm based on local consensus algorithms and ZOO with policy search in the parameter space. ZOO-based RL with policy search in the action space has been proposed in \cite{kumar2020zeroth}. Compared to the action space, the parameter space usually has a higher dimension. However, the work in \cite{kumar2020zeroth} requires the action space to be continuous and leverages the Jacobian of the policy $\pi$ w.r.t. $\theta$. Our RL algorithm is applicable to both continuous and discrete action spaces and even does not require $\pi$ to be differentiable. In addition, our distributed learning framework based on LVFs is compatible with policy search in the action space.

	%In order to optimize the LVF based on the ZOO algorithm, agent $i$ should have access to the value of $\hat{J}_i(\theta)$ when applying a policy $\pi_i(\theta_i)$. Such information can be obtained by inter-agent communications. However, in practice, agent $i$ may not be able to communicate with all the agents $j\in\mathcal{I}_i^L$. Therefore, we design a local consensus algorithm for estimating the LVFs, based on which the distributed RL algorithm will be proposed.

	\subsection{Communication Weights and Distributed RL Design}
	
	We have  shown that agents in the same strongly connected component share the same LVF. Therefore, there are $n$ LVFs to be estimated, where $n$ is the number of maximal SCCs in $\mathcal{G}_{SO}$. Moreover, it is unnecessary for an agent to estimate a LVF that is independent of this agent. For notation simplicity, we use $\mathcal{I}_l^{cl}$ to denote the index set of agents involved in the LVF for the $l$-th cluster, $l\in\mathcal{C}$. As a result, $\mathcal{I}_l^{cl}=\mathcal{I}_i^L$ if $i\in \mathcal{V}_l$. Moreover, we denote by $n_l\triangleq|\mathcal{I}_l^{cl}|$ the number of agents involved in the LVF of cluster $l$. Note that different LVFs for different clusters may involve overlapped agents, that is, it may hold that $\mathcal{I}_{l_1}^{cl}\cap\mathcal{I}_{l_2}^{cl}\neq\varnothing$ for different clusters $l_1$, $l_2$.
	
	Suppose that the communication graph $\mathcal{G}_C=(\mathcal{V},\mathcal{E}_C)$ is available. To make each agent obtain all the individual rewards involved in its LVF, we design a local consensus algorithm based on which the agents involved in each LVF cooperatively estimate the average of their rewards by achieving average consensus. Define $n$ communication weight matrices $C^l\in\mathbb{R}^{N\times N}$, as follows:
	\begin{equation}\label{C matrix}
			C_{ij}^l\left\{
			\begin{aligned}
				&>0,~~\text{if}~ i,j\in\mathcal{I}_l^{cl}, (i,j)\in\mathcal{E}_C;\\
				&=0,~~\text{otherwise},
			\end{aligned}
			\right.~~~~ l\in\mathcal{C},
	\end{equation}
where $\mathcal{C}=\{1,...,n\}$ is the set of indices for clusters.

	We assume that given an initial state, by implementing the global joint policy $\pi(\theta)=(\pi_1(\theta_1),...,\pi_N(\theta_N))$, each agent $i$ is able to obtain reward $r_i(\theta,\xi_i,t)$, at each time step $t=0,...,T_e-1$, where $T_e$ is the number of evolution steps for policy evaluation, $\xi_i$ accounts for the random effects of both the initial states and the state transition of agents involved in agent $i$'s reward, $\mathbb{E}[\xi_i]=0$, and $\mathbb{E}[\xi_i^2]=\sigma_i^2$, which is bounded, $i\in\mathcal{V}$. Then we rewrite the obtained individual value of agent $i$ as  $W_i(\theta,\xi_i)\triangleq\sum_{t=0}^{T_e-1}\gamma^tr_i(\theta,\xi_i,t)=\mathbb{E}[W_i(\theta,\xi_i)]+\xi_i$. The quantity $\xi_i$ can follow any distribution as long as it has a zero mean and a bounded variance. The zero mean assumption is to ensure that agent $i$ is able to evaluate its individual value $W_i(\theta,\xi_i)$ and thereby estimate the gradient accurately, if sufficiently many noisy observations are collected. The boundedness of $\sigma_i$ is to guarantee the boundedness of each noisy observation. Similar assumptions have been made in other RL references, e.g., \cite{vemula2019contrasting}.
	
	We further define $\hat{W}_i(\theta,\xi)=\sum_{j\in\mathcal{I}_i^L}W_j(\theta,\xi_j)$ as the observed LVF value of agent $i$, and define $W(\theta,\xi)=\sum_{i\in\mathcal{V}}W_i(\theta,\xi_i)$ as the observed GVF value.

	The distributed RL algorithm\footnote{In Algorithm \ref{alg}, the transition probability of the MAS is never used. This is consistent with most of model-free RL algorithms in the literature.} is shown in Algorithm \ref{alg}. The consensus algorithm (\ref{local consensus}) is to make each agent $i$ in cluster $\mathcal{V}_l$ estimate $\frac{1}{n_l}\hat{W}_i(\theta^k+\delta u^k,\xi^k)=\frac{1}{n_l}\sum_{j\in\mathcal{I}_i^L}W_j(\theta^k+\delta u^k,\xi_j^k)$, which is the average of the reward sum among the agents involved in the corresponding LVF.

	\begin{algorithm}[htbp]
		\small
		\caption{Distributed RL Algorithm}\label{alg}
		\textbf{Input}: Step-size $\eta$, initial state distribution $\mathcal{D}$, number of learning epochs $K$, number of evolution steps $T_e$ (for policy evaluation), iteration number for consensus seeking $T_c$, initial policy parameter $\theta_0$, smoothing radius $\delta>0$.\\
		\textbf{Output}: $\theta^K$.
		\begin{itemize}
			\item[1.] \textbf{for} $k=0,1,...,K-1$ \textbf{do}
			\item[2.] ~~~~Sample $s_0^k\sim\mathcal{D}$.
			\item[3.] ~~~~\textbf{for} all $i\in\mathcal{V}$ \textbf{do} (Simultaneous Implementation)
			\item[4.] ~~~~~~~Agent $i$ samples $u_{i}^k\sim\mathcal{N}(0,I_{d_i})$, implements policy $\pi_i(\theta_i^k+\delta u_i^k)$ for $t=0, ..., T_e-1$, observes $W_i(\theta^k+\delta u^k,\xi_i^k)$. For $l\in\mathcal{C}$, sets $\mu_i^{kl}(0)\leftarrow W_i(\theta^k+\delta u^k,\xi_i^k)$ if $i\in\mathcal{I}_l^{cl}$, and sets $\mu_i^{kl}(0)\leftarrow0$ otherwise.
			\item[5.] ~~~~~~~~~~\textbf{for}   $v=0,...,T_c-1$ \textbf{do}
			\item[6.] ~~~~~~~~~~~Agent $i$ sends $\mu_i^{kl}(v)$, $l\in\mathcal{C}$ to its neighbors in $\mathcal{G}_C$,  and computes $\mu_i^{kl}(v+1)$ according to the following updating law:
			\begin{equation}\label{local consensus}
					\mu_i^{kl}(v+1)=\sum_{j\in\mathcal{I}_i^C}C_{ij}^l\mu_j^{kl}(v),
			\end{equation}
		where $\mathcal{I}_i^C=\mathcal{N}_i^C\cup\{i\}$, $\mathcal{N}_i^C$ denotes the neighbor set of agent $i$ in the communication graph $\mathcal{G}_C$.
			\item[7.]~~~~~~~~~~\textbf{end for}
			\item[8.]~~~~~~~ Agent $i$ estimates its local gradient 
			\begin{equation}\label{gi}
					g_i(\theta^k,u^k,\xi^k)=\frac{n_{l_i}\mu_i^{kl_i}(T_c)}{\delta}u_i^k,
			\end{equation}
		where $l_i$ denotes the cluster including $i$. Then agent $i$ updates its policy according to
			\begin{equation}
					\theta_i^{k+1}=\theta_i^k+\eta g_i(\theta^k,u^k,\xi^k).
			\end{equation} 
			\item[9.] ~~~~\textbf{end for}
			\item[10.] \textbf{end for}
		\end{itemize}
	\end{algorithm}

	%\begin{algorithm}[htbp]
	%	\small
	%	\caption{Local Value Function Estimation for agent $i$}\label{local value estimation}
	%	\textbf{Input}: iteration number for consensus seeking $T_c$, initial policy parameter $\theta_0$, smoothing radius $\delta>0$.\\
	%	\textbf{Output}: $\theta^K$.\\
	%Agent $i$ sends $\mu_i^{kl}(t)$, $l\in\mathcal{C}$ to its neighbors,  and computes $\mu_i^{kl}(t+1)$ according to the following updating law:
	%		\begin{small}\begin{equation}
			%			\mu_i^{kl}(t+1)=\sum_{j\in\mathcal{V}}C_{ij}^l\mu_j^{kl}(t)+U_i^{kl}, l\in\mathcal{C}.
			%		\end{equation}\end{small}
	%\end{algorithm}	

	To ensure that Algorithm \ref{alg} works efficiently, we make the following assumption on graph $\mathcal{G}_C$.
	
	\begin{assumption}\label{as communication graph}
		The communication graph $\mathcal{G}_C$ is undirected, and the agents specified by $\mathcal{I}_l^{cl}$ form a connected component of $\mathcal{G}_C$ for all $l\in\mathcal{C}$. 
	\end{assumption}
	
	The following lemma gives a sufficient condition for Assumption \ref{as communication graph}.
	
	\begin{lemma}
		Given that graph $\mathcal{G}_C$ is undirected, Assumption \ref{as communication graph} holds if $\mathcal{E}_{SO}\subseteq\mathcal{E}_C$.
	\end{lemma}
	{\it Proof:} Note that for each cluster $l\in\mathcal{C}$, there must exist a path in $\mathcal{G}_{SO}$ from cluster $l$ to any agent  in $\mathcal{I}_l^{cl}$, recall that $\mathcal{G}_C$ is undirected, agents in $\mathcal{I}_l^{cl}$ must be connected in $\mathcal{G}_C$.
	\QEDA
	
	Once a communication graph $\mathcal{G}_C$ satisfying Assumption \ref{as communication graph} is available, we design the communication weights such that the following assumption holds.
	
	\begin{assumption}\label{as C^l}
		$C^l$ is doubly stochastic, i.e.,	$ C^l\mathbf{1}_N=\mathbf{1}_N$ and $\mathbf{1}_N^\top C^l=\mathbf{1}_N^\top$, for all $l\in\mathcal{C}$.
	\end{assumption}
	
	Assumption \ref{as C^l} guarantees that average consensus can be achieved among the agents involved in each LVF. Since one agent may be involved in LVFs of multiple clusters, it may keep multiple different nonzero communication weights for the same communication link. From the definition of $C^l$ in (\ref{C matrix}), $C_{jj'}^l=0$ for all $j\notin\mathcal{I}_i^L$, and $j'\in\mathcal{V}$. Then $\mu_j^{kl}(t)=0$ for $j\notin\mathcal{I}_i^L$ for any $p\geq0$. Moreover, let $C^l_0\in\mathbb{R}^{n_l\times n_l}$ be the principle submatrix of $C^l$ by removing the $j$-th row and column for all $j\notin\mathcal{I}_i^L$, then Assumption \ref{as C^l} implies that $C^l_0$ is doubly stochastic for all $l\in\mathcal{C}$. Define $\rho_l=\|C^l_0-\frac{1}{n_l}\mathbf{1}_{n_l}\mathbf{1}_{n_l}^\top\|$, it has been shown in \cite{xiao2004fast} that under Assumption \ref{as C^l}, we have  $\rho_l\in(0,1)$.
	
	\begin{remark}\label{re convergence rate}
		When graph $\mathcal{G}_{SO}$ is strongly connected, all the agents form one cluster and achieve average consensus during the learning process. Algorithm \ref{alg} then reduces to a global consensus-based distributed RL algorithm. In fact, under any graph $\mathcal{G}_{SO}$, the global consensus-based framework can always solve the distributed RL problem. However, when Assumption \ref{as weakly connected} holds, Algorithm \ref{alg} requires consensus to be achieved among smaller-size groups, therefore exhibiting a faster convergence rate. When the multi-agent network is of large scale, it is possible that the number of agents involved in each LVF is significantly smaller than the total number of agents in the whole network. In such scenarios, Algorithm \ref{alg} converges much faster than the global consensus-based algorithm due to two reasons: (i) the average consensus tasks are performed within smaller-size groups; (ii) the gradient estimation based on the LVF $\hat{J}_i(\theta)$ has a lower variance compared with that based on the GVF $J(\theta)$, see Remark \ref{re variance} for more details.
	\end{remark}
	
	\subsection{Convergence Analysis}
	
	In this subsection, convergence analysis of Algorithm \ref{alg} will be presented. The following assumption is made to guarantee the solvability of the problem (\ref{RL objective}).
	\begin{assumption}\label{as rbound}
		The individual reward of each agent at any time $t$ is uniformly bounded, i.e.,  $r_l\leq r_i(t)\leq r_u$ for all $i\in\mathcal{V}$ and $t\in\mathbb{N}$.
	\end{assumption}

	\begin{lemma}\label{le J*}
		Under Assumption \ref{as rbound}, there exist $J_l$ and $J_u$ such that $J_l\leq J_i(\theta)\leq J_u$ for any $\theta\in\mathbb{R}^d$, $i\in\mathcal{V}$.
	\end{lemma}

	Lemma \ref{le J*} implies that there exists an optimal policy for the RL problem (\ref{maxJ}), which is the premise of solving problem (\ref{maxJ}). Based on Lemma \ref{le J*}, we can bound $\hat{J}_i$ and $J=\sum_{i\in\mathcal{V}}J_i$ by $[\hat{J}_l,\hat{J}_u]$ and $[J_l,J_u]$, respectively. The following lemma bounds the error between the actual LVF and the  expectation of observed LVF.

	\begin{lemma}\label{le J-EW}
		Under Assumption \ref{as rbound}, the following holds for all $l\in\mathcal{C}$ and $i\in\mathcal{V}_l$: 
		\begin{equation}
				|\hat{J}_i(\theta)-\mathbb{E}[\hat{W}_i(\theta,\xi)]|\leq n_l\gamma^{T_e}J_0,
		\end{equation}
	where $J_0=\max\{|J_l|, |J_u|\}=\frac{r_0}{1-\gamma}$, $r_0=\max\{|r_l|,|r_u|\}$, $n_l=|\mathcal{I}_i^L|=|\mathcal{I}_l^{cl}|$ is the number of agents involved in $\hat{J}_i(\theta)$.
	\end{lemma}

	Let $\mu^{kl}=(...,\mu_{j}^{kl},...)_{j\in\mathcal{I}_i^L}^\top\in\mathbb{R}^{|\mathcal{I}_i^L|}$, the following lemma bounds the LVF estimation error.
	
	\begin{lemma}\label{le consensus error}
		Under Assumptions \ref{as weakly connected}, \ref{as communication graph}-\ref{as rbound}, by implementing Algorithm \ref{alg}, the following inequalities hold for any $l\in\mathcal{C}$ and $i\in\mathcal{V}_l$:		
		\begin{equation}\label{Enmu-Jhat}
				|\mathbb{E}[n_l\mu_i^{kl}(T_c)]-\hat{J}_i(\theta^k+\delta u^k)|\leq  E_i,
		\end{equation}
		\begin{equation}\label{Bimu}
				\mathbb{E}_{\xi^k\sim\mathcal{H}}\left[\left[n_l\mu_i^{kl}(T_c)\right]^2\right]\leq B^\mu_l,
		\end{equation}
		where $E_i=\rho_l^{T_c}n_l^2(J_u-J_l+\gamma^{T_e}J_0)+n_l^2\gamma^{T_e}J_0$, $B^\mu_l=n_l^2\left(\sigma_0^2+(1+\gamma^{T_e})^2J_0^2\right)$, $\sigma_0=\max_{i\in\mathcal{V}}\sigma_i$.
	\end{lemma}

	The following lemma bounds the variance of the zeroth-order oracle (\ref{gi}).
	
	\begin{lemma}\label{le variance}
		Under Assumptions \ref{as weakly connected}, \ref{as communication graph}-\ref{as rbound}, for any $i\in\mathcal{V}_l$, it holds that $\mathbb{E}[\|g_i(\theta^k, u^k,\xi^k)\|^2]\leq \frac{B^\mu_l d_i}{\delta^2}$.
	\end{lemma}	
	
	\begin{remark}\label{re variance}(Low Gradient Estimation Variance Induced by LVFs)
		Lemma \ref{le variance} shows that the variance of each local zeroth-order oracle is mainly associated with $n_l$ in $B_l^\mu$, which is the number of agents involved in the LVF for the $l$-th cluster. If the policy evaluation is based on the global reward, the bound of $\mathbb{E}[\|g_i(\theta^k, u^k,\xi^k)\|^2]$ will be $\frac{N^2B^\mu_l d_i}{n_l^2\delta^2}$. When the network is of a large scale, $N$ may be significantly larger than $n_l$. As a result, the variance of the zeroth-order oracle is much higher than that in our case. Therefore, our algorithm  has a significantly improved scalability to large-scale networks.
	\end{remark}

	\begin{theorem}\label{th Lp}
		Under Assumptions \ref{as weakly connected}-\ref{as rbound}, let $\delta=\frac{\epsilon}{L\sqrt{d}}$, $\eta=\frac{\epsilon^{1.5}}{d^{1.5}\sqrt{K}}$. The following statements hold:
		
		(i). $|J^\delta(\theta)-J(\theta)|\leq \epsilon$ for any $\theta\in\mathbb{R}^d$.
		
		(ii). By implementing Algorithm \ref{alg}, if 
		\begin{equation}
				\begin{split}
					&K\geq \frac{d^3B^2}{\epsilon^5}, ~~~~T_e\geq \log_\gamma\frac{\epsilon^{1.5}}{2\sqrt{2}n_0^2LdJ_0},\\
					&T_c\geq \log_{\rho_0}\frac{\epsilon^{1.5}}{2\sqrt{2}n_0^2Ld(J_u-J_l+J_0)},
				\end{split}
		\end{equation}
		then
		\begin{equation}
				\frac{1}{K}\sum_{k=0}^{K-1}\mathbb{E}[\|\nabla_{\theta}J^\delta(\theta^k)\|^2]\leq\epsilon,
		\end{equation}
	where $B=2(NJ_u-J^\delta(\theta^0))+L^4n_0^2(\sigma_0^2+(1+\gamma^{T_e})^2J_0^2)$, $\rho_0=\max_{l\in\mathcal{C}}\rho_l$.
	\end{theorem}	
	
	\begin{remark} (Optimality Analysis)
		Theorem \ref{th Lp} (ii) implies the convergence to a stationary point of $J^\delta(\theta)$, which is the smoothed value function\footnote{The reason why we do not analyze the stationary point of $J(\theta)$ is that we did not assume $J(\theta)$ to be differentiable. Since $J^\delta(\theta)$ is close to $J(\theta)$ (as shown in Theorem \ref{th Lp} (i)), an optimal policy for $J^\delta(\theta)$ will be a near-optimal policy for $J(\theta)$. If we further assume $J(\theta)$ to have a Lipschitz continuous gradient, then the error of convergence to a stationary point of $J(\theta)$ can be obtained by quantifying the error between $\nabla_\theta J^\delta(\theta)$ and $\nabla_\theta J(\theta)$.}. When the MARL problem satisfies ``gradient domination" and the policy parameterization is complete, a stationary point will always be the global optimum \cite{bhandari2019global,agarwal2021theory}. Note that our formulation is general and contains the cases that do not satisfy gradient domination. For example, as a special case of our formulation, the linear optimal distributed control problem has many undesired stationary points \cite{feng2019exponential}.
	\end{remark}
	\begin{remark}(Sample Complexity Analysis)
		According to Theorem \ref{th Lp}, the sample complexity of Algorithm \ref{alg} is $\mathcal{O}(\frac{n_0^4J_0^4}{\epsilon^5})$, which is worse than other ZOO-based algorithms in \cite[Table 1]{zhang2021new}, which is mainly caused by the use of one-point zeroth-order oracles and the mild assumption (non-smoothness and nonconvexity) on the value function. In Section \ref{sec two-point}, we will provide analysis on the advantage of using two-point zeroth-order oracles. Note that the lower bounds of $K$, $T_e$ and $T_c$ are all positively associated with $n_0$, which is the maximal number of agents involved in one LVF. According to the definition of $\mathcal{I}_i^L$ in (\ref{I_i^L}), $n_l$ is determined by the length of the path starting from cluster $l$ in graph $\mathcal{G}_{SO}$. This implies that the convergence rate depends on the maximal length of a path in $\mathcal{G}_{SO}$ and having shorter paths is beneficial for improving sample efficiency and enhancing the convergence rate.
	\end{remark}
	
	When Assumption \ref{as weakly connected} is invalid, Lemmas \ref{le consensus error} and \ref{le variance}, and Theorem \ref{th Lp} still hold. However, the LVF-based method becomes a GVF-based method and no longer exhibits advantages. In this case, there is only one cluster and each path achieves its maximum length, then the distributed RL algorithm becomes centralized and the sample complexity reaches maximum.

	\section{Distributed RL via Truncated Local Value Functions}\label{sec RL TLVF}
	
	Even if Assumption \ref{as weakly connected} holds, the sample complexity of Algorithm \ref{alg} may still be high due to the large size of some SCC or some long paths in $\mathcal{G}_{SO}$. In this section, motivated by \cite{qu2020scalable}, we resolve this issue by further dividing large size SCCs into smaller size SCCs (clusters) and ignoring agents that are far away when designing the LVF for each cluster in the graph. For each cluster, the approximation error turns out to be dependent on the distance between ignored agents and this cluster. Different from \cite{qu2020scalable}, where each agent neglects the effects of other agents that are far away, our design aims to make each cluster neglect its effects on other agents that are far away. Moreover, in our setting, the agents in each cluster estimate their common LVF value via local consensus, whereas in \cite{qu2020scalable}, each agent has its unique LVF, and it was not mentioned how this value can be obtained.
	
	\subsection{Truncated Local Value Function Design}
	
	Different from the aforementioned SCC-based clustering for $\mathcal{G}_{SO}$, now we artificially give a clustering $\mathcal{V}=\cup_{l=1}^{n}\mathcal{V}_l$, where $\mathcal{V}_{l_1}\cap\mathcal{V}_{l_2}=\varnothing$ for distinct $l_1,l_2\leq n$, $\mathcal{V}_l$ still corresponds to a SCC in $\mathcal{G}_{SO}$. However, each cluster may no longer be a maximum SCC. That is, multiple clusters may form a larger SCC of $\mathcal{G}_{SO}$.
	
	Next we define a distance function $D(i,j)$ to describe how many steps are needed for the action of agent $i$ to affect another agent $j\in\mathcal{I}_i^L$. According to Lemma \ref{le GCOR} (i), when $(j,i)\in\mathcal{E}_L$, there is always a path from $i$ to $j$ in $\mathcal{G}_{SOR}$. Let $P(i,j)$ be the length of the shortest path \footnote{The length of a path refers to the number of edges included in this path.}  from vertex $i$ to vertex $j$ in graph $\mathcal{G}_{SOR}$. The distance function is defined as
	\begin{equation}\label{Dij}
			D(i,j)=\left\{
			\begin{array}{lrlrlr}
				0, & i=j,\\
				P(i,j), & (j,i)\in\mathcal{E}_L,\\
				\infty, &  (j,i)\notin\mathcal{E}_L.
			\end{array}
			\right.
	\end{equation}
	We clarify the following facts regarding $D(i,j)$. (i). It may happen that there is a path from $i$ to $j$ in $\mathcal{G}_{SOR}$ but $(j,i)\notin\mathcal{E}_L$, see Fig. \ref{fig counterexample} as an example. Therefore, to exclude $j\notin\mathcal{I}_i^L$, we artificially defined $D(i,j)$ instead of using $P(i,j)$ directly to characterize the inter-agent distance. (ii). The distance function $D(i,j)$ defined here is unidirectional and does not satisfy the symmetry property of the distance in metric space. (iii). Although the artificial SCC clustering is obtained from $\mathcal{G}_{SO}$, the path length $P(i,j)$ is calculated via graph $\mathcal{G}_{SOR}$ because it always contains all the edges from $i$ to any $j\in\mathcal{I}_i^L$. If $\mathcal{G}_{SO}$ is used instead, some agent in $\mathcal{I}_j^R$, $j\in\mathcal{I}_i^L$ may be missing. 
	
	We further define the distance from a cluster $l$ to  an agent $j$ as $D(\mathcal{V}_{l},j)=\min_{i\in\mathcal{V}_{l}}D(i,j)$.
	Denote by $D_l^*=\max_{j\in\mathcal{V}} D(\mathcal{V}_l,j)$ the maximum distance from cluster $l$ to any agent out of this cluster that can be affected by cluster $l$. Since we have defined $n_l=|\mathcal{I}_l^{cl}|$, it is observed that $D_l^*=n_l-|\mathcal{V}_l|$.
	
	Given a cluster $l\in\mathcal{C}$, for any agent $i\in\mathcal{V}_l$, we define the following TLVF:
	\begin{equation}\label{Jtildei}
			\tilde{J}_i^\delta(\theta)=\left\{
			\begin{array}{lrlr}
				\sum_{j\in\mathcal{I}_i^L}J_j^\delta(\theta), & \kappa\geq D_l^*,\\
				\sum_{j\in\mathcal{V}_l^\kappa}J_j^\delta(\theta), &  \kappa<D_l^*,
			\end{array}
			\right.
	\end{equation}
where $\mathcal{V}_l^\kappa=\{j\in\mathcal{V}: D(\mathcal{V}_l,j)\leq\kappa\}$ is the set of agents involved in the TLVF of cluster $l$, $\kappa\in\mathbb{N}_+$ is a pre-specified truncation index describing the maximum distance from each cluster $l$ within which the agents are taken into account in the TLVF of cluster $l$. Similarly, we define $\tilde{J}_i(\theta)=\sum_{j\in\mathcal{I}_i^L}J_j(\theta)$ if $\kappa\geq D_l^*$, and $\tilde{J}_i(\theta)=\sum_{j\in\mathcal{V}_l^\kappa}J_j(\theta)$ otherwise, $\tilde{W}_i(\theta)=\sum_{j\in\mathcal{I}_i^L}W_j(\theta)$ if $\kappa\geq D_l^*$, and $\tilde{W}_i(\theta)=\sum_{j\in\mathcal{V}_l^\kappa}W_j(\theta)$ otherwise.

	The following lemma bounds the error between the local gradients of the TLVF and the GVF.
	\begin{lemma}\label{le nablaJtilde error}
		Under Assumption \ref{as Lip}, given cluster $l$ and agent $i\in\mathcal{V}_l$, the following bound holds for $\tilde{J}^\delta_i(\theta)$:
		\begin{equation}
				\|\nabla_{\theta_i}\tilde{J}_i^\delta(\theta)-\nabla_{\theta_i}J^\delta(\theta)\|\leq \gamma^{\kappa+1}\sum_{j\in\bar{\mathcal{V}}_l^\kappa}L_j\sqrt{dd_i},
		\end{equation}
		where $\bar{\mathcal{V}}_l^\kappa=\{j\in\mathcal{V}: \kappa<D(\mathcal{V}_l,j)<\infty\}=\mathcal{V}_l\setminus\mathcal{V}_l^\kappa$.
	\end{lemma}

	Lemma \ref{le nablaJtilde error} implies that the error between $\nabla_{\theta_i}\tilde{J}_i^\delta(\theta)$ and $\nabla_{\theta_i}J^\delta(\theta)$ exponentially decays with the exponent $\kappa$. Therefore, when $\gamma^{\kappa+1}$ is sufficiently small, by employing $\nabla_{\theta_i}\tilde{J}_i^\delta(\theta)$ in the gradient ascent algorithm, the induced error should be acceptable. This is the fundamental idea of our approach. In next subsection, we will propose the detailed algorithm design and convergence analysis.

	\subsection{Distributed RL with Convergence Analysis}
	
	Next we design a distributed RL algorithm based on the TLVF. It suffices to redesign the communication weights, so that the value of $\tilde{J}_i^\delta(\theta)$ (instead of $\hat{J}^\delta_i(\theta)$) can be estimated for each agent $i\in\mathcal{V}$. For any cluster $l\in\mathcal{C}$, instead of using $\mathcal{I}_l^{cl}$, we set the index set of agents involved in the LVF as $\mathcal{I}_l^\kappa=\mathcal{I}_l^{cl}\cap\mathcal{V}_l^\kappa$.

	\GJ{Define the number of agents involved in each TLVF as $n_l^\kappa$. Then we have $n_l^\kappa\leq n_l$, and the equality holds if $\kappa\geq D_l^*$.}
%	\begin{equation}\label{nbar}
%			n^\kappa_l=\left\{
%			\begin{array}{lrlr}
%				n_l, & \kappa\geq D_l^*,\\
%				|\mathcal{V}_l|+\kappa, & \kappa<D_l^*.
%			\end{array}
%			\right.
%	\end{equation}
%	Recall that $D_l^*=n_l-|\mathcal{V}_l|$, it always holds that $n_l^\kappa\leq n_l$. 
	
	The $l$-th communication weight matrix is then redesigned as
	\begin{equation}\label{C matrix kappa}
			C_{ij}^{l,\kappa}\left\{
			\begin{aligned}
				&>0,~~\text{if}~ i,j\in\mathcal{I}_l^{\kappa}, (i,j)\in\mathcal{E}_C;\\
				&=0,~~\text{otherwise},
			\end{aligned}
			\right.~~~~ l\in\mathcal{C},
	\end{equation}
	where $\mathcal{C}=\{1,...,n\}$.
	
	Similar to Assumptions \ref{as communication graph} and \ref{as C^l}, we make the following two assumptions.
	\begin{assumption}\label{as communication graph kappa}
		The communication graph $\mathcal{G}_C$ is undirected, and the agents specified by $\mathcal{I}_l^{\kappa}$ form a connected component of $\mathcal{G}_C$ for all $l\in\mathcal{C}$. 
	\end{assumption}
	\begin{assumption}\label{as C^l kappa}
		$C^{l,\kappa}$ is doubly stochastic for all $l\in\mathcal{C}$.
	\end{assumption}
	
	Note that Assumption \ref{as communication graph kappa} is milder than Assumption \ref{as communication graph} because $\mathcal{I}_l^\kappa\subseteq\mathcal{I}_l^{cl}$, implying that fewer communication links are needed when the TLVF method is employed. Moreover, when the communication graph $\mathcal{G}_C$ is available, $\kappa$ can be designed to meet Assumption \ref{as C^l kappa} .
	
	The distributed RL algorithm based on TLVFs can be obtained by simply replacing the communication weight matrices $C^l$ with $C^{l,\kappa}$, for all $l\in\mathcal{C}$.
	
	Similar to Lemmas \ref{le J-EW}, \ref{le consensus error} and \ref{le variance}, we have the following results.
	\begin{lemma}\label{le J-EW tr}
		Under Assumption \ref{as rbound}, the following holds for all $l\in\mathcal{C}$ and $i\in\mathcal{V}_l$: 
		\begin{equation}
				|\tilde{J}_i(\theta)-\mathbb{E}[\tilde{W}_i(\theta,\xi)]|\leq n_l^\kappa\gamma^{T_e}J_0.
		\end{equation}
	\end{lemma}
	
	\begin{lemma}\label{le consensus error tr}
		Under Assumptions \ref{as weakly connected}, \ref{as rbound}-\ref{as C^l kappa}, by implementing Algorithm \ref{alg}, the following inequalities hold for any $l\in\mathcal{C}$ and $i\in\mathcal{V}_l$:		
		\begin{equation}\label{Enmu-Jhat tr}
				|\mathbb{E}[n_l\mu_i^{kl}(T_c)]-\tilde{J}_i(\theta^k+\delta u^k)|\leq  E_i^\kappa,
		\end{equation}
		\begin{equation}\label{Bimu tr}
				\mathbb{E}_{\xi^k\sim\mathcal{H}}\left[\left[n_l\mu_i^{kl}(T_c)\right]^2\right]\leq B^\kappa_l,
		\end{equation}
		where $E_i^\kappa=(\rho_l^\kappa)^{T_c}(n_l^\kappa)^2\left(J_u-J_l+\gamma^{T_e}J_0\right)+(n_l^\kappa)^2\gamma^{T_e}J_0$, $B^\kappa_l=(n_l^\kappa)^2(\sigma_0^2+(1+\gamma^{T_e})^2J_0^2)$, $\sigma_0=\max_{i\in\mathcal{V}}\sigma_i$, $\rho_l^\kappa=\|C^{l,\kappa}_0-\frac{1}{n_l^\kappa}\mathbf{1}_{n_l^\kappa}\mathbf{1}_{n_l^\kappa}^\top\|$.
	\end{lemma}
	
	\begin{lemma}\label{le variance tr}
		Under Assumptions \ref{as weakly connected}, \ref{as rbound}-\ref{as C^l kappa}, for any $i\in\mathcal{V}_l$, it holds that $\mathbb{E}[\|g_i(\theta^k, u^k,\xi^k)\|^2]\leq \frac{B^\kappa_l d_i}{\delta^2}$.
	\end{lemma}

	\begin{theorem}\label{th Lp kappa}
		Under Assumptions \ref{as weakly connected}, \ref{as Lip}, \ref{as rbound}-\ref{as C^l kappa}, let $\delta=\frac{\epsilon}{L\sqrt{d}}$, $\eta=\frac{\epsilon^{1.5}}{d^{1.5}\sqrt{K}}$. The following statements hold:
		
		(i). $|J^\delta(\theta)-J(\theta)|\leq \epsilon$ for any $\theta\in\mathbb{R}^d$.
		
		(ii). By implementing Algorithm \ref{alg}, if 
		\begin{equation}
				\begin{split}
					&K\geq \frac{d^3B^2}{\epsilon^5}, ~~~~T_e\geq \log_\gamma\frac{\epsilon^{1.5}}{4(n_0^\kappa)^2LdJ_0}, \\
					&T_c\geq \log_{\rho_0}\frac{\epsilon^{1.5}}{4(n_0^\kappa)^2Ld(J_u-J_l+J_0)},
				\end{split}
		\end{equation}
		then
		\begin{equation}\label{rate}
				\frac{1}{K}\sum_{k=0}^{K-1}\mathbb{E}[\|\nabla_{\theta}J^\delta(\theta^k)\|^2]\leq\epsilon+\gamma^{\kappa+1}\max_{l\in\mathcal{C}}|\bar{\mathcal{V}}_l^\kappa|L_0\sqrt{dd_0},
		\end{equation}
		where $B=2(NJ_u-J^\delta(\theta^0))+(n_0^\kappa)^2(\sigma_0^2+(1+\gamma^{T_e})^2J_0^2)L^4$, $n_0^\kappa=\max_{l\in\mathcal{C}} n_l^\kappa$.
	\end{theorem}

	\begin{remark}\label{re TLVF complexity}(Sample Complexity Analysis)
		The sample complexity provided in Theorem \ref{th Lp kappa} is associated with $n_0^\kappa$, which may be significantly smaller than $n_0$ (depending on the choice of $\kappa$) in Theorem \ref{th Lp}. On the other hand, the convergence error in Theorem \ref{th Lp kappa} has an extra term associated with $\gamma^{\kappa+1}$. Therefore, there is a trade-off when we choose $\kappa$. The greater $\kappa$ is, the smaller the convergence error will be, however, the convergence rate may be decreased. For example, when $\gamma=0.6$, we have $\gamma^{\kappa+1}\leq0.028$ if $\kappa\geq6$. In this case, we can choose $\kappa=6$, implying that each cluster only consider its effects on 6 agents other than this cluster even when $\mathcal{G}_{SO}$ is strongly connected. Therefore, when the network is of a huge scale with long paths in graph $\mathcal{G}_{SO}$, using the TLVFs can further reduce the sample complexity.
	\end{remark}
	
	\section{Variance Reduction by Two-Point Zeroth-Order Oracles} \label{sec two-point}
	
	The two distributed RL algorithms proposed in the last two sections are based on the one-point zeroth-order oracle (\ref{gi}). We observe that Algorithm \ref{alg} is always efficient as long as $g(\theta^k,u^k,\xi^k)$ is an unbiased estimate of $\nabla_\theta J^\delta(\theta^k)$ and $\mathbb{E}[\|g(\theta^k,u^k,\xi^k)\|^2]$ is bounded. Therefore, the two-point feedback oracles proposed in \cite{nesterov2017random} and the residual feedback oracle in \cite{zhang2021new} can also be employed in Algorithm \ref{alg}. In this section, we will give a brief analysis on how the two-point zeroth-order oracle further reduces the gradient estimation variance.
	
	Based on the LVF design in our work, the two-point feedback oracle for each agent $i$ at learning episode $k$ can be obtained as
	\begin{equation}\label{gbar}
			\bar{g}_i(\theta^k, u^k,\xi^k)=\frac{\mu_i^{kl_i}(T_c)-\nu_i^{kl_i}(T_c)}{\delta}n_{l_i}u^k_i,
	\end{equation}
where $\nu_i^{kl_i}(T_c)$ is the approximate estimation of $\hat{W}_i(\theta^k,\zeta^k)/n_{l_i}$ via local consensus.
	
	Define $\hat{L}_i=\sum_{j\in\mathcal{I}_i^L}L_i$, which is a Lipschitz constant of $\hat{W}_i$. Then we can show (\ref{bound on gbar}),
	\begin{figure*}
		
			\begin{equation}\label{bound on gbar}
					\begin{split}
						&\mathbb{E}[\|\bar{g}_i(\theta^k,u^k,\xi^k)\|^2]=\mathbb{E}[\|\frac{\hat{J}_i(\theta^k+\delta u^k)-\hat{J}_i(\theta^k)+n_{l_i}\mu_i^{kl_i}(T_c)-\hat{J}_i(\theta^k+\delta u^k)+\hat{J}_i(\theta^k)-n_{l_i}\nu_i^{kl_i}(T_c)}{\delta}u_i^k\|^2]\\
						&=\mathbb{E}[\|\frac{\hat{J}_i(\theta^k+\delta u^k)-\hat{J}_i(\theta^k)+\mathbb{E}[n_{l_i}\mu_i^{kl_i}(T_c)]-\hat{J}_i(\theta^k+\delta u^k)+\hat{J}_i(\theta^k)-\mathbb{E}[n_{l_i}\nu_i^{kl_i}(T_c)]+\sum_{j\in\mathcal{I}_i^L}(\xi_j^k-\zeta_j^k)}{\delta}u_i^k\|^2]\\
						&\leq 2\mathbb{E}[\hat{L}_i^2\|u^k\|^2\|u_i^k\|^2]+2\left((2E_i)^2+2n_l\sigma_0^2\right)\mathbb{E}[\|u_i^k\|^2/\delta^2]\\
						&\leq2\hat{L}_i^2\left(\mathbb{E}[\|u_i\|^4]+\mathbb{E}[\sum_{j\in\mathcal{V}\setminus\{i\}}\|u_j\|^2]\mathbb{E}[\|u_i\|^2]\right)+4(2E_i^2+n_l\sigma_0^2)d_i/\delta^2\\
						&\leq 2\hat{L}_i^2\left((d_i+4)^2+(d-d_i)d_i\right)+4(2E_i^2+n_l\sigma_0^2)d_i/\delta^2\\
						&=2\hat{L}_i^2(d_id+8d_i+16)+4(2E_i^2+n_l\sigma_0^2)d_i/\delta^2,
					\end{split}
			\end{equation}
	\end{figure*}
	where $\xi=(\xi_1,...,\xi_N)^\top$ and $\zeta=(\zeta_1,...,\zeta_N)^\top$ are the noises in the observations $W_i(\theta^k+\delta u^k,\xi_i^k)$ and $W_i(\theta^k,\zeta^k)$, respectively, i.e., $W_i(\theta^k+\delta u^k,\xi_i^k)=\mathbb{E}[W_i(\theta^k+\delta u^k,\xi_i^k)]+\xi_i$, $W_i(\theta^k,\zeta^k)=\mathbb{E}[W_i(\theta^k,\zeta^k)]+\zeta_i$; the first inequality used the bound $E_i$ in (\ref{Enmu-Jhat}) and the assumptions $\mathbb{E}[\xi_i]=\mathbb{E}[\zeta_i]=0$ and $\mathbb{E}[\xi_i^2]\leq\sigma_0$, $i\in\mathcal{V}$.

	\textbf{Comparisons with one-point feedback.} Note that $E_i$ in (\ref{Enmu-Jhat}) can be arbitrarily small as long as $T_e$ and $T_c$ are sufficiently large. Let us first consider the ideal case where the consensus estimation is perfect $T_c=\infty$ and the observation is exact, i.e., $W_i=\mathbb{E}[W_i]=J_i$ ($T_e=\infty$ and $\xi=0$), then $E_i=0$ and $\sigma_0=0$. As a result, the upper bound of $\mathbb{E}[\|\bar{g}_i(\theta^k,u^k,\xi^k)\|^2]$ is independent of $\delta$, whereas the upper bound of $\mathbb{E}[\|g_i(\theta^k,u^k,\xi^k)\|^2]$ becomes $n_l^2J_0^2d_i/\delta^2$, as shown in Lemma \ref{le variance}. This implies that the variance of the two-point zeroth-order oracle is independent of the reward value and the maximum path length $n_0$, thus is more scalable than the one-point feedback. Now we consider a more practical scenario where both the consensus estimations and the observations are inexact. For convenience of analysis, we consider $\delta>0$ as an infinitesimal quantity and neglect terms in the upper bounds independent of the network scale. Then we have $\mathbb{E}[\|\bar{g}_i(\theta^k,u^k,\xi^k)\|^2]=\mathcal{O}(n_ld_i/\delta^2)$. In Lemma \ref{le variance}, we showed that the variance bound for the zeroth-order oracle with one-point feedback is $\mathcal{O}(n_l^2d_i/\delta^2)$. Therefore, when $\delta>0$ is small enough, the two-point zeroth-order oracle still outperforms the one-point feedback scheme in terms of lower variance and faster convergence speed.

	\section{Simulation Results}\label{sec sim}
	
	In this section, we present two examples, where the first one shows the results of applying Algorithm \ref{alg} with the communication weight matrices (\ref{C matrix}) to the resource allocation problem in Example \ref{ex warehouse}, and the second one shows the results of applying Algorithm \ref{alg} with the communication weight matrices (\ref{C matrix kappa}) to a large-scale network scenario.
	
	\begin{example}\label{ex 9warehouses}
		We employ the distributed RL with LVFs in (\ref{Jhati}) to solve the problem in Example \ref{ex warehouse}. To seek the optimal policy $\pi_i(o_i)$ for agent $i$ to determine its action $\{b_{ij}\}_{j\in\mathcal{N}_i^{S+}}$, we adopt the following parameterization for the policy function:
		\begin{equation}
				b_{ij}=\frac{\exp(-z_{ij})}{\sum_{j\in\mathcal{I}_i}\exp(-z_{ij})},
		\end{equation}
		where $z_{ij}$ is approximated by radial basis functions:
		\begin{equation}
				z_{ij}=\sum_{k=1}^{n_c}\|o_i-c_{ik}\|^2\theta_{ij}(k),
		\end{equation}
		$c_{ik}=(\hat{c}_{ik}^\top,\bar{c}_{ik})\in\mathbb{R}^{|\mathcal{I}_i^O|+1}$ is the center of the $k$-th feature for agent $i$, here $\hat{c}_{ik}\in\mathbb{R}^{|\mathcal{I}_i^O|}$ and $\bar{c}_{ik}$ are set according to the ranges of $m_{\mathcal{I}_i}$ and $d_i$, respectively, such that $c_{ik}$, $k=1,...,n_c$ are approximately evenly distributed in the range of $o_i$.
		
		Set $m_i(0)=1+\chi_i$ for all $i=1,...,9$, $\chi_i$ and $w_i$ are both set as random variables following the Guassian distribution with mean 0 amd variance 0.01 truncated to $[-0.01,0.01]$, the number of evolution steps $T_e=10$, and the number of learning epochs $K=1500$, $y_i(t)-d_i(t)=0.5\sin t$. The communication graph $\mathcal{G}_C=(\mathcal{V},\mathcal{E}_C)$ is set as $\mathcal{G}_C=\mathcal{G}_{SO}\cup\mathcal{G}_{SO}^\top$, which satisfies Assumption \ref{as communication graph}. Let $G_C$ be the 0-1weighted matrix of graph $\mathcal{G}_C$, that is, $G_C(i,j)=1$ if $(i,j)\in\mathcal{E}_C$ and $G_C(i,j)=0$ otherwise. Let $d_i^C=\sum_{(j,i)\in\mathcal{E}_C}G_C(i,j)$. The communication weights are set as Metropolis weights \cite{xiao2005scheme}:
		\begin{equation}\label{Clij}
				C^l_{ij}=\left\{
				\begin{aligned}
					&\frac{1}{1+\max\{d_i^C,d_j^C\}},~~\text{if}~i\neq j,~ i,j\in\mathcal{I}_l^{cl}, (i,j)\in\mathcal{E}_C;\\
					&1-\sum_{j\neq i}C^l_{ij},~~~~~~~~~~~~~\text{if} ~i=j;\\
					&0,~~~~~~~~~~~~~~~~~~~~~~~~~\text{otherwise},
				\end{aligned}
				\right.~~~~ 
		\end{equation}
	where $l\in\mathcal{C}$.
		
		By further setting the consensus iteration number $T_c=10$, $\eta=0.01$, and $\delta=2$, Fig. \ref{Fig.4} (left) depicts the evolution of the observed values of the GVF by implementing 4 different RL algorithms. The two boundaries of the shaded area are obtained by running each RL algorithm for 10 times and taking the upper bound and lower bound of $W(\theta^k,\xi^*)$ in each learning episode. Here $\xi^*$ denotes the specific noise generated in the simulation, and is different in different learning processes. In each time of implementation, one perturbation vector $u^k$ is sampled and used for all the 4 algorithms during each learning episode $k$. The centralized algorithm is the zeroth-order optimization algorithm based on global value evaluation, while the distributed algorithm is based on local value evaluation (Algorithm \ref{alg}). The distributed two-point feedback algorithm is Algorithm \ref{alg} with $g_i(\theta^k,u^k,\xi^k)$ replaced by $\bar{g}_i(\theta^k,u^k,\xi^k)$ in (\ref{gbar}). We observe that the distributed algorithms are always faster than the centralized algorithms. Fig. \ref{Fig.4} (middle) and Fig. \ref{Fig.4} (right) show the comparison of centralized and distributed one-point feedback algorithms, and the comparison of centralized and distributed two-point feedback algorithms, respectively. From these two figures, it is clear that the distributed algorithms always exhibit lower variances in contrast to the centralized algorithms. This implies that policy evaluation based on LVFs is more robust than that based on the GVF.
		
		\begin{figure*}%[htbp]
			%\vspace{-0.8cm}
			\centering
			\includegraphics[width=1\textwidth]{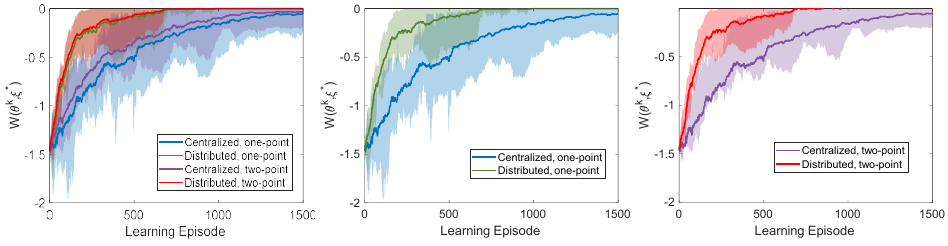}%
			\caption{\textbf{(Left)} Comparison of different algorithms for 9 warehouses;  \textbf{(Middle)} centralized and distributed algorithms under zeroth-order oracles with one-point feedback; \textbf{(Right)} centralized and distributed algorithms under zeroth-order oracles with two-point feedback. The observed GVF value $W(\theta^k,\xi^*)=\sum_{i\in\mathcal{V}}W_i(\theta^k,\xi_i^*)$ is employed as the performance metric. The boundaries of the shaded area are obtained by running each RL algorithm for 10 times and taking the upper bound and lower bound of $W(\theta^k,\xi^*)$ in each learning episode.}
			\label{Fig.4}
		\end{figure*} 
		
	\end{example}
	
	\begin{example}\label{ex 100agents}
		Next, in a setting similar to Example \ref{ex warehouse}, we consider an extendable example with $N$ warehouses. By regarding 1 and $N+1$ as the same warehouse, we set
		%\begin{small}
		\begin{equation*}
			\mathcal{E}_S=\mathcal{E}_O=\{(i,j)\in\mathcal{V}^2: |i-j|=1, i=2k-1, k\in\mathbb{N}\}.
		\end{equation*}
		%\end{small}
		The reward graph is set as $\mathcal{G}_R=\mathcal{G}_S^\top$. According to the definitions introduced in Subsection \ref{subsec LVF design}, we have $\mathcal{I}_i^L=\{i,i+1,i+2,i-1,i-2\}$ if $i$ is odd, $\mathcal{I}_i^L=\{i,i+1,i-1\}$ if $i$ is even, for all $i\in\mathcal{V}$.
		
		The communication graph $\mathcal{G}_C=(\mathcal{V},\mathcal{E}_C)$ is set as $\mathcal{G}_C=\mathcal{G}_L$, implying that each agent can estimate its LVF value without using the local consensus algorithm. The learning iteration step is set as $\eta=0.05$. Other parameter settings are the same as those in Example \ref{ex 9warehouses}. By implementing four different RL algorithms, Fig. \ref{fig comparison} shows the results for $N=20$, $N=40$, and $N=80$, respectively. Observe that the convergence time for distributed algorithms remain almost invariant for networks with different scales, whereas the centralized algorithms converge much slower when the network scale is increased. Moreover, the two-point oracle always outperforms the one-point oracle in terms of lower variance and faster convergence. These observations are consistent with our analysis in Remark \ref{re convergence rate} and Section \ref{sec two-point}.
		
		\begin{figure*}%[htbp]
			\centering
			\includegraphics[width=1\textwidth]{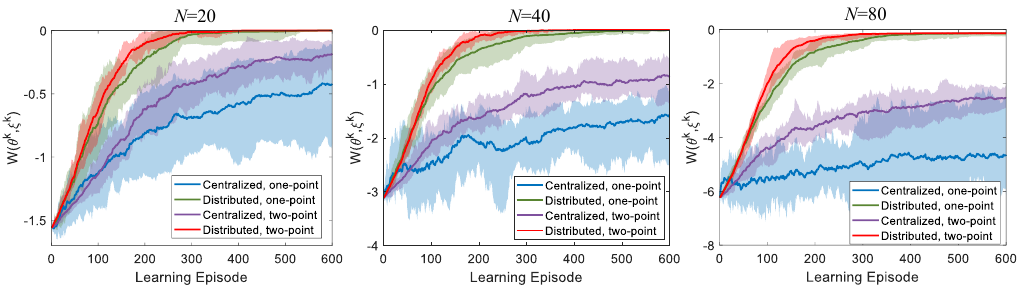}%
			\caption{Comparison of different RL algorithms for Example 3 with $N=20$, $N=40$, and $N=80$, respectively.} 
			\label{fig comparison}
		\end{figure*}

		\begin{figure}%[htbp]
			%\vspace{-0.6cm}
			\centering
			\includegraphics[width=0.45\textwidth]{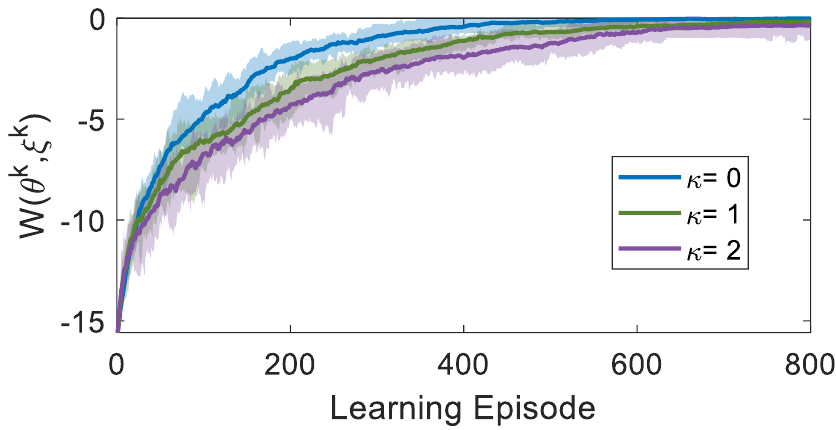}%
			\caption{Comparison of Algorithm \ref{alg} based on TLVFs with $\kappa=0$, 1, and 2, respectively, for Example \ref{ex TLVF}.} 
			\label{fig 100agents}
		\end{figure} 
	\end{example}
	
	\begin{example}\label{ex TLVF}
		Now we consider $N=100$ warehouses with connected undirected state graph and observation graph,
		\begin{equation*}
			\mathcal{E}_S=\mathcal{E}_O=\{(i,j)\in\mathcal{V}^2: |i-j|=1\},
		\end{equation*}
		where warehouse 1 is also viewed as warehouse $N+1$.
		
		The edge set of graph $\mathcal{G}_R=(\mathcal{V},\mathcal{E}_R)$ is set as $\mathcal{E}_R=\varnothing$, implying that the individual reward of each agent only depends on its own state and action. In this case, the learning graph $\mathcal{G}_L$ is complete because $i\stackrel{\mathcal{E}_{SO}}{\longrightarrow} j$ for any $i,j\in\mathcal{V}$. Then Algorithm \ref{alg} with the LVF setting in (\ref{Jhati}) becomes  a centralized algorithm. Hence, we employ the TLVF defined in (\ref{Jtildei}). The communication graph is set as $\mathcal{G}_C=\mathcal{G}_S$. By setting $\mathcal{V}_l=\{4l-3,...,4l\}$, $l=1,...,25$, and choosing same parameters $m_i(0)$, $T_e$, $z_i(t)$ as those in Example \ref{ex 9warehouses}, the simulation results for $\kappa=0$, 1 and 2 are shown in Fig. \ref{fig 100agents}. We observe that the distributed RL algorithm with $\kappa=0$ gains the lowest variance and fastest convergence rate. This means that in this example, the TLVF approximation error does not harm the improved performance of our RL algorithm. Moreover, the smaller $\kappa$ is, the faster the algorithm converges, which is consistent with the analysis in Remark \ref{re TLVF complexity}. %Note that according to the analysis in Example 3, each cluster corresponds to one agent, and the maximum distance between a cluster $i$ and an agent affected by this cluster is 2 (in $\mathcal{G}_{SOR}$). Therefore, we only considered cases with $\kappa\leq2$. 
	\end{example}
	
	\section{Conclusions}\label{sec conclusion}
	
	We have recognized three graphs inherently embedded in MARL, namely, the state graph, observation graph, and reward graph. A connection between these three graphs and the learning graph was established, based on which we proposed our distributed RL algorithm via LVFs and derived conditions on the communication graph required in RL. It was shown that the LVFs constructed based on the aforementioned 3 graphs are able to play the same role as the GVF in gradient estimation. To adapt our algorithm to MARL with general graphs, we have designed TLVFs associated with an artificially specified index. The choice of this index is a trade-off between variance reduction and gradient approximation errors. Simulation examples have shown that our proposed algorithms with LVFs or TLVFs significantly outperform RL algorithms based on GVF, especially for large-scale MARL problems.
	
	The RL algorithms proposed in this work are policy gradient algorithms based on ZOO, which are general, but may be not the best choice for specific applications. In the future, we are interested in exploring how our graph-theoretic approach can be combined with other RL techniques to facilitate learning for large-scale network systems.

	\section{Appendix A: Proofs of Lemmas and Theorems}\label{sec: appendix A}
	
	{\it Proof of Lemma \ref{le GCOR}.} (i). Suppose that $\mathcal{I}_i^L=\mathcal{V}$ for all $i\in\mathcal{V}$. Since $\mathcal{G}_{SOR}$ has $n>1$ SCCs, there must exist distinct $i,j\in\mathcal{V}$ such that $j$ is not reachable from $i$ in $\mathcal{G}_{SOR}$. However, $\mathcal{I}_i^L=\mathcal{V}$ implies that there exists $k\in\mathcal{V}$ such that $j\in\mathcal{I}_k^{R+}$, and $k\in\mathcal{R}_i^{SO}$, implying that $j$ is reachable from $i$ in $\mathcal{G}_{SO}\cup\mathcal{G}_R$, which is a contradiction. 
	
	(ii). Suppose $\mathcal{G}_{SO}$ only has 1 SCC. According to (\ref{R_i^SO}), $\mathcal{R}_i^{SO}=\mathcal{V}$ for any $i\in\mathcal{V}$. This implies that $\mathcal{I}_i^L=\mathcal{V}$ for any $i\in\mathcal{V}$, which contradicts with Assumption \ref{as weakly connected}.
	\QEDA
	
	{\it Proof of Lemma \ref{le lg=gg}.}
	Define 
	\begin{small}\begin{equation}
			\bar{J}_i(\theta)=J(\theta)-\hat{J}_i(\theta)=\sum_{j\in\mathcal{V}\setminus\mathcal{I}_i^L}\mathbb{E}_{s_0\sim\mathcal{D}}V_j^{\pi(\theta)}(s_0). 
	\end{equation}\end{small}
	Next we show that $\bar{J}_i(\theta)$ is independent of $\theta_i$.
	
	Let $\mathcal{R}^{SO-}_i=\{j\in\mathcal{V}: j\stackrel{\mathcal{E}_{SO}}{\longrightarrow} i\}\cup\{i\}$ be the set of vertices in graph $\mathcal{G}_{SO}$ that can reach $i$ and vertex $i$. Note that for each agent $j$, its action at time $t$, i.e., $a_j(t)$, is only affected by the partial observation $o_j(t)$, the current state $s_j(t)$, and policy $\theta_j$. Therefore, there exists a function $f_j:\mathcal{O}_j\times\mathbb{R}^{d_j}\rightarrow P(\mathcal{A}_j)$ such that 
	\begin{small}\begin{equation}\label{ajt}
			\begin{split}
				a_j(t)\sim f_j(o_j(t),\theta_j)=f_j(\{s_{k}(t)\}_{k\in\mathcal{I}_j^O}, \theta_j).
			\end{split}		
	\end{equation}\end{small}
	Similarly, according to the definition of $\mathcal{P}_i$, there exists another function $h_j:\Pi_{k\in\mathcal{I}_j^S}\mathcal{S}_k\times\Pi_{k\in\mathcal{I}_j^S}\mathcal{A}_k\rightarrow P(\mathcal{S}_j)$ such that
	\begin{small}\begin{equation}\label{sjt}
			s_j(t)\sim h_j( \{s_k(t-1)\}_{k\in\mathcal{I}_j^S}, \{a_k(t-1)\}_{k\in\mathcal{I}_j^S}),
	\end{equation}\end{small}together with (\ref{ajt}), we have
	\begin{small}\begin{equation}\label{sjtsk}
			s_j(t)\sim h_j( \{s_k(t-1)\}_{k\in\mathcal{I}_j^{SO}}, \{\theta_l\}_{l\in\mathcal{I}_j^S}),
	\end{equation}\end{small}and
	\begin{small}\begin{equation}\label{ajt2}
			\begin{split}
				a_j(t)\sim f_j(\{h_k(\{s_{l_1}(t-1)\}_{l_1\in\mathcal{I}_k^{SO}},\{\theta_{l_2}\}_{l_2\in\mathcal{I}_k^S})\}_{k\in\mathcal{I}_j^O}, \theta_j),
			\end{split}		
	\end{equation}\end{small}where $\mathcal{I}_j^{SO}=\mathcal{I}_j^S\cup\mathcal{I}_j^O$.
	
	According to (\ref{sjtsk}) and  (\ref{ajt2}), we conclude that $\{s_j(t),a_j(t)\}$ is affected by $\theta_l$ only if $l\in\mathcal{R}_j^{SO-}$. As a result, $\{s_k(t),a_k(t)\}_{k\in\mathcal{I}_j^R}$ is affected by $\theta_l$ only if $l\in\cup_{k\in\mathcal{I}_j^R}\mathcal{R}_k^{SO-}\triangleq A_j$.
	
	Next we show once $j\notin\mathcal{I}_i^L$, it must hold that $i\notin A_j$, i.e., $\theta_i$ will not affect $\{s_k(t),a_k(t)\}_{k\in\mathcal{I}_j^R}$. By the definition in (\ref{I_i^L}), $j\notin\mathcal{I}_i^L$ implies that $\mathcal{I}_j^R\cap\mathcal{R}_i^{SO}=\varnothing$. That is, there are no vertices in $\mathcal{I}_j^R$ that are reachable from vertex $i$ in graph $\mathcal{G}_{SO}$. As a result, $i\notin A_j$.
	
	Then we conclude that  $\theta_i$ never influences $r_j(s_{\mathcal{I}_i^R}(t),a_{\mathcal{I}_i^R}(t))$ for any $t\geq0$ if $j\notin\mathcal{I}_i^L$. Therefore, $\bar{J}_i(\theta)$ is independent of $\theta_i$.
	
	Proof for (i):	it has been shown in \cite{nesterov2017random} that
	\begin{small}\begin{equation}\label{phi}
			\nabla_\theta J^\delta(\theta)=\frac{1}{\delta\varphi}\int_{\mathbb{R}^d} J(\theta+\delta u)e^{-\frac12\|u\|^2}udu,
	\end{equation}\end{small}where $\varphi=\int_{\mathbb{R}^d}e^{-\frac12\|u\|^2}du$.
	Define \begin{small}\begin{equation}\label{Phii}
			\Phi^i=(\mathbf{0}_{d_i\times d_1},..., I_{d_i},...,\mathbf{0}_{d_i\times d_N})\in\mathbb{R}^{d_i\times d},
	\end{equation}\end{small}then $u_i=\Phi^iu$. It follows that
	\begin{small}%\begin{equation}
		\begin{align}
			\nabla_{\theta_i}J^\delta(\theta)&=\Phi^i\nabla_\theta J^\delta(\theta)\nonumber\\
			&=\frac{1}{\delta\varphi}\int_{\mathbb{R}^d} J(\theta+\delta u)e^{-\frac12\|u\|^2}u_idu\nonumber\\
			&=\frac{1}{\delta\varphi}\int_{\mathbb{R}^d} \hat{J}_i(\theta+\delta u)e^{-\frac12\|u\|^2}u_idu\\
			&~~+\frac{1}{\delta\varphi}\int_{\mathbb{R}^d} \bar{J}_i(\theta+\delta u)e^{-\frac12\|u\|^2}u_idu.\nonumber
		\end{align}
		%\end{equation}
	\end{small}Let $\bar{\theta}_i=(...,\theta_j^\top,...)^\top_{j\neq i}\in\mathbb{R}^{d-d_i}$, $\bar{u}_i=(...,u_j^\top,...)^\top_{j\neq i}\in\mathbb{R}^{d-d_i}$. Since we have proved that $\bar{J}_i(\theta+\delta u)$ is independent of $u_i$, the following holds:
	\begin{equation*}
		\begin{split}
			&\int_{\mathbb{R}^d} \bar{J}_i(\theta+\delta u)e^{-\frac12\|u\|^2}u_idu\\&
			=\int_{\mathbb{R}^{d-d_i}} \bar{J}_i(\bar{\theta}_i+\delta\bar{u}_i)e^{-\frac12\|\bar{u}_i\|^2}d\bar{u}_i\int_{\mathbb{R}^{d_i}} e^{-\frac12\|u_i\|^2}u_idu_i=0.
		\end{split}	
	\end{equation*}
	Therefore, 
	\begin{small}\begin{equation}
			\nabla_{\theta_i}J^\delta(\theta)=\frac{1}{\delta\varphi}\int_{\mathbb{R}^d} \hat{J}_i(\theta+\delta u)e^{-\frac12\|u\|^2}u_idu=\nabla_{\theta_i}\hat{J}_i^\delta(\theta).
	\end{equation}\end{small}
	
	Proof for (ii): differentiability of  $J_i(\theta)$ for all $i\in\mathcal{V}$ implies that $\hat{J}_i(\theta)$ for all $i\in\mathcal{V}$ and $J(\theta)$ are differentiable as well. Since $\bar{J}_i(\theta)$ is independent of $\theta_i$, we have
	\begin{small}\begin{equation}
			\nabla_{\theta_i}J(\theta)=\nabla_{\theta_i}(\hat{J}_i(\theta)+\bar{J}_i(\theta))=\nabla_{\theta_i}\hat{J}_i(\theta).
	\end{equation}\end{small}
	
	This completes the proof.
	\QEDA

	{\it Proof of Lemma \ref{le J*}.} Given any policy $\pi(\theta)$, it holds that
	\begin{small}\begin{equation}
			\sum_{t=0}^\infty\gamma^tr_i(s_i(t),a_i(t))\leq \sum_{t=0}^\infty\gamma^t r_u=\frac{1}{1-\gamma}r_u\triangleq J_u.
	\end{equation}\end{small}Similarly, it can be shown that $J_l=\frac{1}{1-\gamma}r_l$.
	\QEDA
	
	{\it Proof of Lemma \ref{le J-EW}.} From the definition of $J_i(\theta)$ and $W_i(\theta,\xi_i)$, we have
	\begin{small}\begin{equation}
			\begin{split}
				\left|J_i(\theta)-\mathbb{E}[W_i(\theta,\xi)]\right|&=\left|\mathbb{E}\left[\sum_{t=T_e}^{\infty}\gamma^tr_i(s_i(t),a_i(t))\right]\right|\\&\leq \sum_{t=T_e}^\infty\gamma^tr_0=\frac{\gamma^{T_e}}{1-\gamma}r_0.
			\end{split}
	\end{equation}\end{small}Then we have
	\begin{small}\begin{equation*}
			|\hat{J}_i(\theta)-\mathbb{E}[\hat{W}_i(\theta,\xi)]|\leq\sum_{j\in\mathcal{I}_i^L}\left|J_j(\theta)-\mathbb{E}[W_j(\theta,\xi)]\right|\leq\frac{n_l\gamma^{T_e}}{1-\gamma}r_0.
	\end{equation*}\end{small}The proof is completed.
	\QEDA
	
	{\it Proof of Lemma \ref{le consensus error}.} Note that the following holds for any step $v$:
	\begin{small}\begin{equation}\label{mup+1=mup}
			\sum_{j\in\mathcal{I}_i^L}\mu_j^{kl}(v+1)=\mathbf{1}_{n_l}^\top\mu^{kl}(v+1)=\mathbf{1}_{n_l}^\top C^l_0\mu^{kl}(v)=\mathbf{1}_{n_l}^\top\mu^{kl}(v),
	\end{equation}\end{small}where the last equality holds because $C_0^l$ is doubly stochastic. It follows that $\mathbf{1}_{n_l}^\top\mu^{kl}(v)=\hat{W}_i(\theta^k+\delta u^k,\xi_i^k)$ for any $v\in[0,T_c]$.
	
	Next we evaluate the estimation error. The following holds:
	\begin{small}\begin{equation}\label{mu(t+1)tomu0}
			\begin{split}
				&n_l\mu^{kl}(v)-\mathbf{1}_{n_l}\hat{W}_i(\theta^k+\delta u^k,\xi_i^k)\\
				&=n_l\mu^{kl}(v)-\mathbf{1}_{n_l}\mathbf{1}_{n_l}^\top\mu^{kl}(v)\\
				&=n_lC^l_0\mu^{kl}(v-1)-\mathbf{1}_{n_l}\mathbf{1}_{n_l}^\top\mu^{kl}(v-1)\\
				&=(C^l_0-\frac{1}{n_l}\mathbf{1}_{n_l}\mathbf{1}_{n_l}^\top) (n_l\mu^{kl}(v-1)-\mathbf{1}_{n_l}\mathbf{1}_{n_l}^\top\mu^{kl}(v-1))\\
				&=(C^l_0-\frac{1}{n_l}\mathbf{1}_{n_l}\mathbf{1}_{n_l}^\top)^{v}(n_l\mu^{kl}(0)-\mathbf{1}_{n_l}\mathbf{1}_{n_l}^\top\mu^{kl}(0))\\
				&=(C^l_0-\frac{1}{n_l}\mathbf{1}_{n_l}\mathbf{1}_{n_l}^\top)^{v}(n_l\mu^{kl}(0)-\mathbf{1}_{n_l}\hat{W}_i(\theta^k+\delta u^k,\xi_i^k)),
			\end{split}
	\end{equation}\end{small}where the second equality used (\ref{mup+1=mup}) and the third equality holds because $(C^l_0-\frac{1}{n_l}\mathbf{1}_{n_l}\mathbf{1}_{n_l}^\top)\mathbf{1}_{n_l}\mathbf{1}_{n_l}^\top\mu^{kl}(t)=0$.
	
	As a result, for any $v\in [0,T_c]$, we have
	%\begin{small}\begin{equation}\label{rho^m+1}
	\begin{align}\label{rho^m+1}
		&\left\|\mathbb{E}\left[n_l\mu^{kl}(v)\right]-\mathbf{1}_{n_l}\hat{J}_i(\theta^k+\delta u^k)\right\|\nonumber\\
		&\leq\left\|\mathbb{E}\left[n_l\mu^{kl}(v)-\mathbf{1}_{n_l}\hat{W}_i(\theta^k+\delta u^k,\xi_i^k)\right]\right\|\nonumber\\
		&+\left\|\mathbf{1}_{n_l}\hat{J}_i(\theta^k+\delta u^k)-\mathbb{E}\left[\mathbf{1}_{n_l}\hat{W}_i(\theta^k+\delta u^k,\xi_i^k)\right]\right\|\nonumber\\
		&\leq \rho_l^{v}\left\|\mathbb{E}\left[n_l\mu^{kl}(0)-\mathbf{1}_{n_l}\hat{W}_i(\theta^k+\delta u^k,\xi_i^k)\right]\right\|+n_l^2\gamma^{T_e}J_0\nonumber\\
		&\leq \rho_l^{v}\left\|\mathbb{E}\left[n_l\mu^{kl}(0)-\mathbf{1}_{n_l}\hat{J}_i(\theta^k+\delta u^k)\right]\right\|+(\rho_l^{v}+1)n_l^2\gamma^{T_e}J_0\nonumber\\
		&\leq \rho_l^{v}n_l^2(J_u-J_l)+(\rho_l^{v}+1)n_l^2\gamma^{T_e}J_0\nonumber\\
		&=\rho_l^{v}n_l^2\left(J_u-J_l+\gamma^{T_e}J_0\right)+n_l^2\gamma^{T_e}J_0,
	\end{align}
	%\end{equation}\end{small}
	where the second inequality used (\ref{mu(t+1)tomu0}) and Lemma \ref{le J-EW}, the third inequality used Lemma \ref{le J-EW} again, and the last inequality used the uniform bound of $J_i$ and the fact that $\mathbb{E}[\mu^{kl}_i(0)]=\mathbb{E}[W_i(\theta^k+\delta u^k,\xi_i^k)]\leq J_i^\delta(\theta^k)$.

	Due to Assumption \ref{as C^l}, we have $\min_{j\in\mathcal{I}_i^L}W_j(\theta^k+\delta u^k,\xi_i^k)\leq\mu_i^{kl}(T_c)\leq \max_{j\in\mathcal{I}_i^L}W_j(\theta^k+\delta u^k,\xi_i^k)$. Let $i_0=\arg\max_{j\in\mathcal{I}_i^L}|W_j(\theta^k+\delta u^k,\xi_i^k)|$. Then we have 
	%\begin{small}\begin{equation}
	\begin{align}
		&\mathbb{E}_{\xi^k\sim\mathcal{H}}\left[\left[n_l\mu_i^{kl}(T_c)\right]^2\right]\leq n_l^2\mathbb{E}_{\xi^k\sim\mathcal{H}}\left[W_{i_0}^2(\theta^k+\delta u^k,\xi^k)\right]\nonumber\\
		&= n_l^2\mathbb{E}_{\xi^k\sim\mathcal{H}}[(\xi^k_{i_0})^2]+n_l^2(\mathbb{E}_{\xi^k\sim\mathcal{H}}[W_{i_0}(\theta^k+\delta u^k,\xi^k)])^2\nonumber\\
		&\leq n_l^2\sigma_0^2+n_l^2(J_{i_0}(\theta^k+\delta u^k)+\gamma^{T_e}J_0)^2\nonumber\\
		&\leq n_l^2(\sigma_0^2+(1+\gamma^{T_e})^2J_0^2).
	\end{align}
	%\end{equation}\end{small}
	
	The proof is completed.
	\QEDA

	{\it Proof of Lemma \ref{le variance}.} According to (\ref{gi}), we have
	%\begin{small}\begin{equation}
	\begin{align}
		\mathbb{E}[\|g_i(\theta^k, &u^k,\xi^k)\|^2]=\frac{1}{\delta^2}\mathbb{E}_{u_i^k}\left[\mathbb{E}_{\xi^k\sim\mathcal{H}}\left[\left(n_{l}\mu_i^{kl}(T_c)\right)^2\right]\|u_i^k\|^2\right]\nonumber\\
		&\leq\frac{B^\mu_l}{\delta^2}\mathbb{E}\left[\|u^k_i\|^2\right]\nonumber\\
		&= \frac{B^\mu_l}{\delta^2\varphi}\int_{\mathbb{R}^{d_i}}\|u_i^k\|^2e^{-\frac12\|u_i^k\|^2}du_i^k\int_{\mathbb{R}^{d-d_i}}e^{-\frac12\|v\|^2}dv\nonumber\\
		&\leq\frac{B^\mu_l}{\delta^2\varphi}d_i\int_{\mathbb{R}^{d_i}}e^{-\frac12\|u_i^k\|^2}du_i^k\int_{\mathbb{R}^{d-d_i}}e^{-\frac12\|v\|^2}dv\nonumber\\
		&=\frac{B^\mu_l d_i}{\delta^2},
	\end{align}
	%\end{equation}\end{small}
	where $\varphi$ is defined in (\ref{phi}), the first inequality used (\ref{Bimu}), and the second inequality holds because $\int_{\mathbb{R}^{d_i}}\|u_i^k\|^2e^{-\frac12\|u_i^k\|^2}du_i^k\leq d_i\int_{\mathbb{R}^{d_i}}e^{-\frac12\|u_i^k\|^2}du_i^k$, which has been proved in \cite[Lemma 1]{nesterov2017random}.
	\QEDA

	{\it Proof of Theorem \ref{th Lp}.}
	Statement (i) can be obtained by using the Lipschitz continuity of $J(\theta)$. The details have been shown in \cite[Theorem 1]{nesterov2017random}. 
	
	Now we prove statement (ii). According to Assumption \ref{as Lip} and \cite[Lemma 2]{nesterov2017random}, the gradient of $J^\delta(\theta)$ is $\sqrt{d}L/r$-Lipschitz continuous. Let $g(\theta^k)=(g_1^\top(\theta^k),...,g_N^\top(\theta^k))^\top\in\mathbb{R}^d$, the following holds:
	\begin{small}
		\begin{multline}
			|J^\delta(\theta^{k+1})- J^\delta(\theta^k)-\langle\nabla_{\theta}J^\delta(\theta^k),\eta g(\theta^k)\rangle|\\\leq\frac{\sqrt{d}L}{2r}\eta^2\|g(\theta^k,u^k,\xi^k)\|^2,
	\end{multline}\end{small}which implies that
	\begin{small}\begin{multline}\label{nablaJetag}
			\langle\nabla_{\theta}J^\delta(\theta^k),\eta g(\theta^k)\rangle\\
			\leq J^\delta(\theta^{k+1})- J^\delta(\theta^k)+\frac{\sqrt{d}L}{2r}\eta^2\|g(\theta^k,u^k,\xi^k)\|^2.
	\end{multline}\end{small}
	
	Lemma \ref{le variance} implies that 
	\begin{small}\begin{equation}\label{||g||^2}
			\begin{split}
				\mathbb{E}[\|g(\theta^k, u^k,\xi^k)\|^2]&=\sum_{i=1}^N \mathbb{E}[\|g_i(\theta^k, u^k,\xi^k)\|^2]\\ &\leq \sum_{i=1}^NB^\mu_l d_i/\delta^2\leq B^\mu_0 d/\delta^2,
			\end{split}
	\end{equation}\end{small}where $B_0^\mu=\max_{l\in\mathcal{C}} B^\mu_l=n_0^2(\sigma_0^2+(1+\gamma^{T_e})^2J_0^2)$.
	
	Moreover, Lemma \ref{le consensus error} implies that
	%\begin{small}\begin{equation}
	\begin{align}\label{EnJigi1}
		&\mathbb{E}\left[\langle\nabla_{\theta_i}J^\delta(\theta^k),g_i(\theta^k,u^k,\xi^k)\rangle\right]\nonumber\\
		&= \mathbb{E}[\|\nabla_{\theta_i}J^\delta(\theta^k)\|^2]\nonumber\\
		&~~~+\mathbb{E}[\langle\nabla_{\theta_i}J^\delta (\theta^k), n_l\mu_i^{kl}(T_c)u^k_i/\delta-\hat{J}_i(\theta^k+\delta u^k)u^k_i/\delta\rangle]\nonumber\\
		&\geq \mathbb{E}[\|\nabla_{\theta_i}J(\theta^k)\|^2] -\frac{1}{2}\mathbb{E}\left[\|\nabla_{\theta_i}J(\theta^k)\|^2+E_i^2\|u_i^k\|^2/\delta^2\right]\nonumber\\
		&=\frac12\mathbb{E}[\|\nabla_{\theta_i}J(\theta^k)\|^2]-\frac{E_i^2d_i}{2\delta^2},
	\end{align}
	%\end{equation}\end{small}
	where the first equality used Lemma \ref{le lg=gg} and $\nabla_{\theta_i}\hat{J}^\delta(\theta^k)=\mathbb{E}[\hat{J}_i(\theta^k+\delta u^k)u^k_i/\delta]$, the inequality used (\ref{Enmu-Jhat}). Summing (\ref{EnJigi1}) over $i$ from $0$ to $N$ yields
	\begin{small}\begin{equation}\label{nablaJg}
			\mathbb{E}[\langle\nabla_{\theta}J^\delta(\theta^k),g(\theta^k,u^k,\xi^k)\rangle]\geq \frac12\mathbb{E}[\|\nabla_{\theta}J(\theta^k)\|^2]-\frac{E_0^2d}{2\delta^2},
	\end{equation}\end{small}where $E_0=\max_{i\in\mathcal{V}}E_i$.

	Combining (\ref{nablaJg}) and (\ref{nablaJetag}), and taking expectation on both sides, we obtain
	\begin{small}
		\begin{align}\label{etanabla}
			&\frac12\eta\mathbb{E}[\|\nabla_{\theta}J^\delta(\theta^k)\|^2]-\eta\frac{E_0^2d}{2\delta^2}\nonumber\\
			&\leq \mathbb{E}[J^\delta(\theta^{k+1})- J^\delta(\theta^k)]+\frac{\sqrt{d}L}{2\delta}\eta^2\mathbb{E}[\|g(\theta^k,u^k,\xi^k)\|^2]\nonumber\\
			&\leq \mathbb{E}[J^\delta(\theta^{k+1})- J^\delta(\theta^k)] +\frac{LB^\mu_0 d^{1.5}}{2\delta^3}\eta^2,
		\end{align}
	\end{small}where the second inequality employed (\ref{||g||^2}).
	
	Note that under the conditions on $T_c$, we have
	\begin{small}\begin{equation}\label{E0B}
			E_0\leq \frac{\epsilon^{1.5}}{\sqrt{2}Ld}.
	\end{equation}\end{small}
	
	Summing (\ref{etanabla}) over $k$ from 0 to $K-1$ and dividing both sides by $K$, yields
	\begin{small}
		\begin{align}
			&\frac{1}{K}\sum_{k=0}^{K-1}\mathbb{E}[\|\nabla_{\theta}J^\delta(\theta^k)\|^2]\nonumber\\ &\leq\frac{2}{\eta}\left[\frac1K\left(\mathbb{E}[J^\delta(\theta^K)]-J^\delta(\theta^0)\right)+\frac{LB^\mu_0 d^{1.5}}{2\delta^3}\eta^2\right]+\frac{E_0^2d}{\delta^2}\nonumber\\
			&\leq\frac{2}{\eta}\left[\frac{1}{K}(NJ_u-J^\delta(\theta^0))+B_0^\mu\frac{Ld^{1.5}}{2\delta^3}\eta^2\right]+\frac{\epsilon}{2}\nonumber\\
			&\leq\frac{d^{1.5}}{\epsilon^{1.5} \sqrt{K}}\left[ 2(NJ_u-J^\delta(\theta^0))+L^4B_0^\mu\right]+\frac{\epsilon}{2},\label{averagenabla}
		\end{align}
	\end{small}where the second inequality used (\ref{E0B}), the last inequality used the conditions on $\delta$ and $\eta$.
	The proof is completed.	
	\QEDA
	
	{\it Proof of Lemma \ref{le nablaJtilde error}.}  When $\kappa\geq D^*_l$, (\ref{Jtildei}) implies $\tilde{J}_i^\delta(\theta)=\hat{J}_i^\delta(\theta)$. By Lemma \ref{le lg=gg},  $\|\nabla_{\theta_i}\tilde{J}_i^\delta(\theta)-\nabla_{\theta_i}J^\delta(\theta)\|=0$. Next we analyze the other case.
	
	Let $r_i^\theta(s_i(t),a_i(t))$ be the individual reward of agent $i$ at time $t$ under the global policy $\pi(\theta)$. Due to Lemma \ref{le lg=gg}, it suffices to analyze $\|\nabla_{\theta_i}\tilde{J}_i^\delta-\nabla_{\theta_i}\hat{J}_i^\delta\|$.
	
	Let $J_{i,\kappa}^\delta(\theta)=\sum_{j\in\bar{\mathcal{V}}_l^\kappa}\sum_{t=0}^\kappa\gamma^tr_j(t)(s_j(t),a_j(t))$, and $\bar{J}^\delta_{i,\kappa}(\theta)=\sum_{j\in\bar{\mathcal{V}}_l^\kappa}\sum_{t=\kappa+1}^\infty\gamma^tr_j(t)(s_j(t),a_j(t))$. Then 
	\begin{small}\begin{equation}
			\tilde{J}^\delta_i(\theta)-\hat{J}_i^\delta(\theta)=J_{i,\kappa}^\delta(\theta)+\bar{J}^\delta_{i,\kappa}(\theta).
	\end{equation}\end{small}Notice that if $D(\mathcal{V}_l,j)>\kappa$, then the reward of each agent $j\in\bar{\mathcal{V}}_l^\kappa$ is not affected by cluster $\mathcal{V}_l$ at any time step $t\leq \kappa$. Therefore, $\nabla_{\theta_i}J_{i,\kappa}^\delta(\theta)=0$, which leads to (\ref{gJbar-gJhat}),
	\begin{figure*}
		\begin{small}
			\begin{equation}\label{gJbar-gJhat}
				\begin{split}
					&\|\nabla_{\theta_i}\tilde{J}_i^\delta(\theta)-\nabla_{\theta_i}\hat{J}_i^\delta(\theta)\|=\|\nabla_{\theta_i}\bar{J}^\delta_{i,\kappa}(\theta)\|\\
					&=\Bigg\|\mathbb{E}\left[\sum_{j\in\bar{\mathcal{V}}_l^\kappa}\frac{\sum_{t=\kappa+1}^\infty\gamma^tr_j^{\theta+\delta u}(s_j(t),a_j(t))-\sum_{t=\kappa+1}^\infty\gamma^tr_j^{\theta}(s_j(t),a_j(t))}{\delta}u_i\right]\Bigg\|\\
					&\leq\sum_{j\in\bar{\mathcal{V}}_l^\kappa}\mathbb{E}_{u\sim\mathcal{N}(0,I_d)}\left[\|\mathbb{E}_{s\sim\mathcal{D}}\sum_{t=\kappa+1}^\infty\gamma^t\frac{r_j^{\theta+\delta u}(s_j(t),a_j(t))-r_j^\theta(s_j(t),a_j(t))}{\delta}u_i\Big|s_0=s\|\right]\\
					&=\gamma^{\kappa+1} \sum_{j\in\bar{\mathcal{V}}_l^\kappa} \mathbb{E}_{u\sim\mathcal{N}(0,I_d)}\left[\|\mathbb{E}_{s\sim\mathcal{D}}\left[\frac{V^{\pi(\theta+\delta u)}_j(s_{j,\kappa+1})-V^{\pi(\theta)}_j(s_{j,\kappa+1})}{\delta}u_i\Big|s_0=s\right]\|\right] \\
					&\leq\gamma^{\kappa+1}\sum_{j\in\bar{\mathcal{V}}_l^\kappa}L_j\mathbb{E}[\|u\|\|u_i\|]\\
					&\leq\gamma^{\kappa+1}\sum_{j\in\bar{\mathcal{V}}_l^\kappa}L_j(\mathbb{E}[\|u\|^2])^{1/2}(\mathbb{E}[\|u_i\|^2])^{1/2}\\
					&\leq\gamma^{\kappa+1}\sum_{j\in\bar{\mathcal{V}}_l^\kappa}L_j\sqrt{dd_i},
				\end{split}
			\end{equation}
		\end{small}
	\end{figure*}
	where  the second equality used the two-point feedback zeroth-order oracle \cite{nesterov2017random}, the third equality used the definition of $V_i^{\pi(\theta)}(s)$, the second inequality used Assumption \ref{as Lip}, the third inequality used Holder's inequality. The proof is completed.
	\QEDA

	{\it Proof of Theorem \ref{th Lp kappa}.} Here we only show the different part compared with the proof of Theorem \ref{th Lp}.
	\begin{small}
		%\begin{small}\begin{equation}\label{EnJigi}
		\begin{align}\label{EnJigi}
			&\mathbb{E}\left[\langle\nabla_{\theta_i}J^\delta(\theta^k),g_i(\theta^k,u^k,\xi^k)\rangle\right]= \mathbb{E}[\|\nabla_{\theta_i}J^\delta(\theta^k)\|^2]\nonumber\\
			&+\mathbb{E}[\langle\nabla_{\theta_i}J^\delta (\theta^k), n_l\mu_i^{kl}(T_c)u^k_i/\delta-\tilde{J}_i(\theta^k+\delta u^k)u^k_i/\delta\rangle]\nonumber\\
			&+\mathbb{E}[\langle\nabla_{\theta_i}J^\delta (\theta^k),\nabla_{\theta_i}\tilde{J}_i(\theta^k)-\nabla_{\theta_i}\hat{J}_i(\theta^k)\rangle]\nonumber\\
			&\geq \mathbb{E}[\|\nabla_{\theta_i}J(\theta^k)\|^2] -\frac{1}{2}\mathbb{E}\left[\|\nabla_{\theta_i}J(\theta^k)\|^2\right]-(E_i^\kappa)^2\|u_i^k\|^2/\delta^2-\frac{d_i}{\delta^2}A_i^2\nonumber\\
			&=\frac12\mathbb{E}[\|\nabla_{\theta_i}J(\theta^k)\|^2]-\frac{(E_i^\kappa)^2d_i}{\delta^2}-\frac{d_i}{\delta^2}A_i^2,
		\end{align}
	\end{small}where $A_i=\gamma^{\kappa+1}\sum_{j\in\bar{\mathcal{V}}_l^\kappa}L_j\sqrt{dd_i}$, the first equality used $\mathbb{E}_{u^k_i}[\tilde{J}_i(\theta^k+\delta u^k)u^k_i/\delta]=\nabla_{\theta_i}\tilde{J}_i(\theta^k)$ and $\mathbb{E}_{u^k_i}[\hat{J}_i(\theta^k+\delta u^k)u^k_i/\delta]=\nabla_{\theta_i}\hat{J}_i(\theta^k)$, the inequality used $ab\geq -\frac14a^2-b^2$.
	Then we have
	\begin{small}\begin{equation*}\label{nablaJg kappa}
			\mathbb{E}[\langle\nabla_{\theta}J^\delta(\theta^k),g(\theta^k,u^k,\xi^k)\rangle]\geq \frac12\mathbb{E}[\|\nabla_{\theta}J(\theta^k)\|^2]-\frac{(E_0^\kappa)^2d}{\delta^2}-\frac{d}{\delta^2}A_0^2,
	\end{equation*}\end{small}where $E_0^\kappa=\max_{i\in\mathcal{V}}E_i^\kappa$, $A_0=\max_{i\in\mathcal{V}}A_i=\gamma^{\kappa+1}\max_{l\in\mathcal{C}}|\bar{\mathcal{V}}_l^\kappa|L_0\sqrt{dd_0}$.
	
	It follows that
	\begin{small}
		\begin{align}\label{etanabla tr}
			&\frac12\eta\mathbb{E}[\|\nabla_{\theta}J^\delta(\theta^k)\|^2]-\eta\frac{(E_0^\kappa)^2d}{2\delta^2}-\frac{d}{\delta^2}A_0^2\nonumber\\
			&\leq \mathbb{E}[J^\delta(\theta^{k+1})- J^\delta(\theta^k)]+\frac{\sqrt{d}L}{2\delta}\eta^2\mathbb{E}[\|g(\theta^k,u^k,\xi^k)\|^2]\nonumber\\
			&\leq \mathbb{E}[J^\delta(\theta^{k+1})- J^\delta(\theta^k)] +\frac{LB^\kappa_0 d^{1.5}}{2\delta^3}\eta^2,
		\end{align}
	\end{small}where $B_0^\kappa=(n_l^\kappa)^2(\sigma_0^2+J_0^2)$.

	Therefore, the following holds:
	\begin{small}
		\begin{align}
			&\frac{1}{K}\sum_{k=0}^{K-1}\mathbb{E}[\|\nabla_{\theta}J^\delta(\theta^k)\|^2]\nonumber\\ &\leq\frac{2}{\eta}\left[\frac1K\left(\mathbb{E}[J^\delta(\theta^K)]-J^\delta(\theta^0)\right)+\frac{LB^\kappa_0 d^{1.5}}{2\delta^3}\eta^2\right]+\frac{2E_0^2d}{\delta^2}+\frac{2d}{\delta^2}A_0^2\nonumber\\
			&\leq\frac{2}{\eta}\left[\frac{1}{K}(NJ_u-J^\delta(\theta^0))+(n_l^\kappa)^2(\sigma_0^2+J_0^2)\frac{Ld^{1.5}}{2\delta^3}\eta^2\right]+\frac{\epsilon}{2}+\frac{2d}{\delta^2}A_0^2\nonumber\\
			&\leq\frac{d^{1.5}}{\epsilon^{1.5} \sqrt{K}}\left[ 2(NJ_u-J^\delta(\theta^0))+(n_l^\kappa)^2(\sigma_0^2+J_0^2)L^4\right]+\frac{\epsilon}{2}+\frac{2d}{\delta^2}A_0^2.\label{averagenabla kappa}
		\end{align}
	\end{small}
	The proof is completed.
	\QEDA

	\section{Appendix B: Properties of Cluster-Wise Graphs}\label{sec: appendix B}

	In this appendix, we will analyze the relationships among graphs $\mathcal{G}_X$, $X\in\{S, O, R, L\}$ from the cluster-wise perspective.
	
\begin{figure}%[h]
		\vspace{-0.6cm}
		\centering
		\includegraphics[width=7cm]{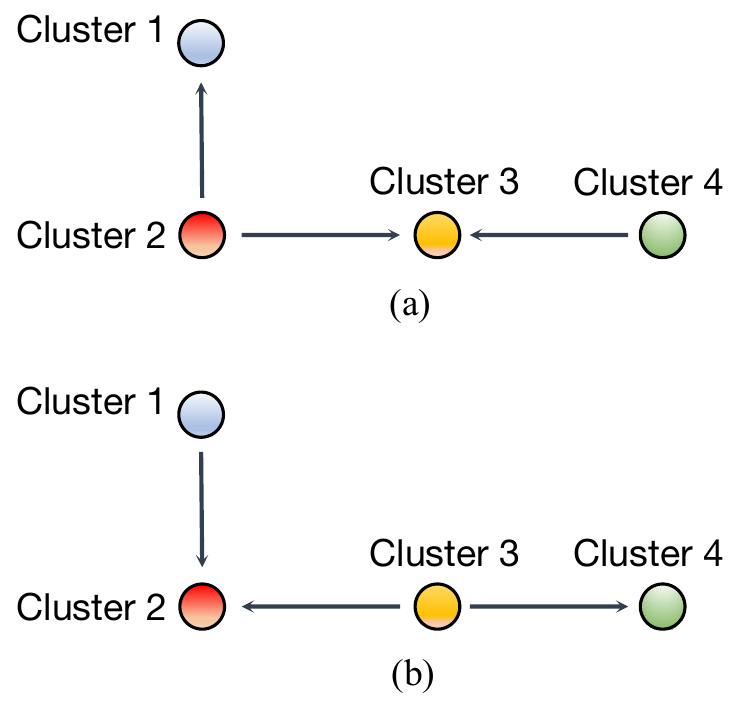}
		\caption{(a). The cluster-wise learning graph $\mathcal{G}_L^{cl}$. (b). The cluster-wise graph $\mathcal{G}_{SO}^{cl}$ or $\mathcal{G}_{SOR}^{cl }$.} \label{fig cluster graphs}
	\end{figure}

	Inspired by the observations in Subsection \ref{subsec: learning graph}, the graph $\mathcal{G}_L$ in Fig. \ref{fig learning graph} can be interpreted from a cluster perspective. By regarding each cluster\footnote{Please note that the clustering in this paper is only conducted once for graph $\mathcal{G}_{SO}$. The clusters discussed in other graphs still correspond to SCCs in graph $\mathcal{G}_{SO}$.} (corresponding to a maximal SCC in $\mathcal{G}_{SO}$) as a node, and adding a directional edge $(l_1,l_2)$ between any pair of nodes $l_1, l_2\in\mathcal{C}$ as long as there is at least one edge from cluster $l_1$ to cluster $l_2$ in $\mathcal{G}_L$, we define the cluster-wise graph of $\mathcal{G}_L$ as the graph $\mathcal{G}_L^{cl}$ in Fig. \ref{fig cluster graphs} (a). Similarly, we define cluster-wise graphs for $\mathcal{G}_{SO}$ and $\mathcal{G}_{SOR}$ as $\mathcal{G}_{SO}^{cl}$ and $\mathcal{G}_{SOR}^{cl}$, respectively. In Example \ref{ex warehouse}, due to the setting that there are no edges between different clusters in $\mathcal{G}_R$, it holds that $\mathcal{G}_{SO}^{cl}=\mathcal{G}_{SOR}^{cl}$, as shown in Fig. \ref{fig cluster graphs} (b). Note that $\mathcal{G}_{SO}\neq\mathcal{G}_{SOR}$ since $\mathcal{E}_R\not\subseteq \mathcal{E}_{SO}$.
	The cluster-wise graph $\mathcal{G}^{cl}$ for a graph $\mathcal{G}$ is constructed by regarding each cluster as one node, and adding one edge between two nodes if there is an edge between two agents belonging to these two clusters in $\mathcal{G}$, respectively. Note that a SCC in $\mathcal{G}_{SO}$ remains to be a SCC in $\mathcal{G}_L$ and $\mathcal{G}_{SOR}$. Therefore, an edge from cluster $l_1$ to cluster $l_2$ in $\mathcal{G}^{cl}$ always implies that any vertex $j\in\mathcal{V}_{l_2}$ is reachable from any vertex $i\in\mathcal{V}_{l_1}$ in the corresponding node-wise graph $\mathcal{G}$. Based on this fact, we give a specific result regarding the relationship between $\mathcal{G}_L^{cl}$, $\mathcal{G}_{SO}^{cl}$ and $\mathcal{G}_{SOR}^{cl}$ as below.
	\begin{comment}
		However, the specific relationship between a graph and its cluster-wise version may differ, depending on the structure of the graph, see the following details.
		\begin{itemize}
			\item $(l_1,l_2)\in\mathcal{E}_{SO}^{cl}$ implies that there exists one edge in $\mathcal{G}_{SO}$ from SCC cluster $\mathcal{V}_{l_1}$ to SCC cluster $\mathcal{V}_{l_2}$, in other words, any vertex in cluster $\mathcal{V}_{l_2}$ is reachable from any vertex in cluster $\mathcal{V}_{l_1}$.
			\item $(l_1,l_2)\in\mathcal{E}_L^{cl}$ implies that there exists at least one edge from some vertex in cluster $\mathcal{V}_{l_1}$ to some vertex in cluster $\mathcal{V}_{l_2}$.
		\end{itemize} 
	\end{comment}

	\begin{lemma}\label{le GL and GCO}
		Given $\mathcal{G}_S$, $\mathcal{G}_O$, $\mathcal{G}_R$ and the induced $\mathcal{G}_L$, the following statements are true:
		
		(i) $(\mathcal{E}_{SO}^{cl})^\top\subseteq\mathcal{E}_L^{cl}\subseteq(\mathcal{E}_{SOR}^{cl})^\top$;
		
		(ii) $(\mathcal{G}_{SO}^{cl})^\top=\mathcal{G}_L^{cl}=(\mathcal{G}_{SOR}^{cl})^\top$ if $\mathcal{E}_R\subseteq\mathcal{E}_{SO}$;
		
		(iii) $(\mathcal{G}_{SO}^{cl})^\top=\mathcal{G}_L^{cl}=(\mathcal{G}_{SOR}^{cl})^\top$ if and only if $i\stackrel{\mathcal{E}_{SO}}{\longrightarrow} j$ for any $ (i,j)\in\mathcal{E}_R$.
	\end{lemma}
	{\it Proof.} (i). Given $(l_1,l_2)\in\mathcal{E}_{SO}^{cl}$, there must hold that $ i\stackrel{\mathcal{E}_{SO}}{\longrightarrow} j$ for any $i\in\mathcal{V}_{l_1}$ and $j\in\mathcal{V}_{l_2}$. Due to the definition of $\mathcal{G}_L$, we have $j\in\mathcal{I}_i^L$ (i.e., $(j,i)\in\mathcal{E}_L$) for any $i\in\mathcal{V}_{l_1}$ and $j\in\mathcal{V}_{l_2}$, implying that $(l_1,l_2)\in(\mathcal{E}_L^{cl})^\top$. Therefore, $\mathcal{E}_{SO}^{cl}\subseteq(\mathcal{E}_L^{cl})^\top$. It follows that $(\mathcal{E}_{SO}^{cl})^\top\subseteq\mathcal{E}_L^{cl}$.
	
	On the other hand, for any $(l_1,l_2)\in\mathcal{E}_L^{cl}$, we have $(i,j)\in\mathcal{E}_L$ for some $i\in\mathcal{V}_{l_1}$ and $j\in\mathcal{V}_{l_2}$. According to Lemma \ref{le ELtoECOR}, $j\stackrel{\mathcal{E}_{SOR}}{\longrightarrow} i$. Hence, $(l_1,l_2)\in(\mathcal{E}_{SOR}^{cl})^\top$.
	
	(ii). By the virtue of statement (i), it suffices to show that $\mathcal{G}_{SO}^{cl}=\mathcal{G}_{SOR}^{cl}$ if $\mathcal{E}_R\subseteq\mathcal{E}_{SO}$, which is true due to the definition of $\mathcal{G}_{SOR}$.
	
	(iii). Sufficiency. The condition implies that the reachability between any two vertices in $\mathcal{G}_{SO}$ is the same as that in $\mathcal{G}_{SOR}$. Therefore, $\mathcal{E}_{SO}^{cl}=\mathcal{E}_{SOR}^{cl}$. 
	
	Necessity. Suppose that there exists an edge $(i,j)\in\mathcal{E}_R$ such that $j$ is not reachable from $i$ in $\mathcal{G}_{SO}$. Then $i$ and $j$ must belong to two different clusters $l_1$ and $l_2$, respectively. It follows that $(l_1,l_2)\in\mathcal{E}_{SOR}$ and $(l_1,l_2)\notin\mathcal{E}_{SO}$, which contradicts with $\mathcal{G}_{SO}^{cl}=\mathcal{G}_{SOR}^{cl}$.
	%Note that $\mathcal{G}_{SO}^{cl}=\mathcal{G}_{SOR}^{cl}$ is equivalent to the statement that there is a path from $i$ to $j$ in $\mathcal{G}_{SO}$ if and only if there is a path from $i$ to $j$ in $\mathcal{G}_{SOR}$, which is equivalent to the statement that adding the edges of $\mathcal{E}_R^\top$ to $\mathcal{G}_{SO}$ does not change the reachability between any two nodes in $\mathcal{G}_{SO}$. This holds if and only if $j$ is reachable from $i$ for any $(i,j)\in\mathcal{E}_R$.
	\QEDA
	
	In the existing literature of MARL, it is common to see the assumption that $\mathcal{I}_i^R=\{i\}$. In this scenario, $\mathcal{E}_R=\varnothing\subseteq\mathcal{E}_{SO}$, therefore it always holds that $\mathcal{G}_L^{cl}=(\mathcal{G}_{SO}^{cl})^\top=(\mathcal{G}_{SOR}^{cl})^\top$.

	\bibliography{Reference} % bibliography data in report.bib

\begin{thebibliography}{10}
\providecommand{\url}[1]{#1}
\csname url@samestyle\endcsname
\providecommand{\newblock}{\relax}
\providecommand{\bibinfo}[2]{#2}
\providecommand{\BIBentrySTDinterwordspacing}{\spaceskip=0pt\relax}
\providecommand{\BIBentryALTinterwordstretchfactor}{4}
\providecommand{\BIBentryALTinterwordspacing}{\spaceskip=\fontdimen2\font plus
\BIBentryALTinterwordstretchfactor\fontdimen3\font minus
  \fontdimen4\font\relax}
\providecommand{\BIBforeignlanguage}[2]{{%
\expandafter\ifx\csname l@#1\endcsname\relax
\typeout{** WARNING: IEEEtran.bst: No hyphenation pattern has been}%
\typeout{** loaded for the language `#1'. Using the pattern for}%
\typeout{** the default language instead.}%
\else
\language=\csname l@#1\endcsname
\fi
#2}}
\providecommand{\BIBdecl}{\relax}
\BIBdecl

\bibitem{sutton2018reinforcement}
R.~S. Sutton and A.~G. Barto, \emph{Reinforcement learning: An
  introduction}.\hskip 1em plus 0.5em minus 0.4em\relax MIT press, 2018.

\bibitem{silver2017mastering}
D.~Silver, J.~Schrittwieser, K.~Simonyan, I.~Antonoglou, A.~Huang, A.~Guez,
  T.~Hubert, L.~Baker, M.~Lai, A.~Bolton \emph{et~al.}, ``Mastering the game of
  go without human knowledge,'' \emph{nature}, vol. 550, no. 7676, pp.
  354--359, 2017.

\bibitem{polydoros2017survey}
A.~S. Polydoros and L.~Nalpantidis, ``Survey of model-based reinforcement
  learning: Applications on robotics,'' \emph{Journal of Intelligent \& Robotic
  Systems}, vol.~86, no.~2, pp. 153--173, 2017.

\bibitem{mukherjee2021scalable}
S.~Mukherjee, A.~Chakrabortty, H.~Bai, A.~Darvishi, and B.~Fardanesh,
  ``Scalable designs for reinforcement learning-based wide-area damping
  control,'' \emph{IEEE Transactions on Smart Grid}, vol.~12, no.~3, pp.
  2389--2401, 2021.

\bibitem{oroojlooyjadid2019review}
A.~OroojlooyJadid and D.~Hajinezhad, ``A review of cooperative multi-agent deep
  reinforcement learning,'' \emph{arXiv preprint arXiv:1908.03963}, 2019.

\bibitem{gronauer2021multi}
S.~Gronauer and K.~Diepold, ``Multi-agent deep reinforcement learning: a
  survey,'' \emph{Artificial Intelligence Review}, pp. 1--49, 2021.

\bibitem{zhang2021multi}
K.~Zhang, Z.~Yang, and T.~Ba{\c{s}}ar, ``Multi-agent reinforcement learning: A
  selective overview of theories and algorithms,'' \emph{Handbook of
  Reinforcement Learning and Control}, pp. 321--384, 2021.

\bibitem{qu2020scalable}
G.~Qu, A.~Wierman, and N.~Li, ``Scalable reinforcement learning of localized
  policies for multi-agent networked systems,'' in \emph{Learning for Dynamics
  and Control}.\hskip 1em plus 0.5em minus 0.4em\relax PMLR, 2020, pp.
  256--266.

\bibitem{lin2021multi}
Y.~Lin, G.~Qu, L.~Huang, and A.~Wierman, ``Multi-agent reinforcement learning
  in stochastic networked systems,'' in \emph{Thirty-Fifth Conference on Neural
  Information Processing Systems}, 2021.

\bibitem{omidshafiei2017deep}
S.~Omidshafiei, J.~Pazis, C.~Amato, J.~P. How, and J.~Vian, ``Deep
  decentralized multi-task multi-agent reinforcement learning under partial
  observability,'' in \emph{International Conference on Machine
  Learning}.\hskip 1em plus 0.5em minus 0.4em\relax PMLR, 2017, pp. 2681--2690.

\bibitem{nayak2023scalable}
S.~Nayak, K.~Choi, W.~Ding, S.~Dolan, K.~Gopalakrishnan, and H.~Balakrishnan,
  ``Scalable multi-agent reinforcement learning through intelligent information
  aggregation,'' in \emph{International Conference on Machine Learning}.\hskip
  1em plus 0.5em minus 0.4em\relax PMLR, 2023, pp. 25\,817--25\,833.

\bibitem{zhang2020cooperative}
Y.~Zhang and M.~M. Zavlanos, ``Cooperative multi-agent reinforcement learning
  with partial observations,'' \emph{IEEE Transactions on Automatic Control},
  2023, doi: 10.1109/TAC.2023.3288025.

\bibitem{zhang2018fully}
K.~Zhang, Z.~Yang, H.~Liu, T.~Zhang, and T.~Basar, ``Fully decentralized
  multi-agent reinforcement learning with networked agents,'' in
  \emph{International Conference on Machine Learning}.\hskip 1em plus 0.5em
  minus 0.4em\relax PMLR, 2018, pp. 5872--5881.

\bibitem{guestrin2002coordinated}
C.~Guestrin, M.~Lagoudakis, and R.~Parr, ``Coordinated reinforcement
  learning,'' in \emph{ICML}, vol.~2.\hskip 1em plus 0.5em minus 0.4em\relax
  Citeseer, 2002, pp. 227--234.

\bibitem{kok2006collaborative}
J.~R. Kok and N.~Vlassis, ``Collaborative multiagent reinforcement learning by
  payoff propagation,'' \emph{Journal of Machine Learning Research}, vol.~7,
  pp. 1789--1828, 2006.

\bibitem{jing2021asynchronous}
G.~Jing, H.~Bai, J.~George, A.~Chakrabortty, and P.~K. Sharma, ``Asynchronous
  distributed reinforcement learning for {LQR} control via zeroth-order block
  coordinate descent,'' \emph{arXiv preprint arXiv:2107.12416}, 2021.

\bibitem{li2021distributed}
Y.~Li, Y.~Tang, R.~Zhang, and N.~Li, ``Distributed reinforcement learning for
  decentralized linear quadratic control: A derivative-free policy optimization
  approach,'' \emph{IEEE Transactions on Automatic Control}, 2021.

\bibitem{gorges2019distributed}
D.~G{\"o}rges, ``Distributed adaptive linear quadratic control using
  distributed reinforcement learning,'' \emph{IFAC-PapersOnLine}, vol.~52,
  no.~11, pp. 218--223, 2019.

\bibitem{Jingtcns}
G.~Jing, H.~Bai, J.~George, and A.~Chakrabortty, ``Model-free optimal control
  of linear multi-agent systems via decomposition and hierarchical
  approximation,'' \emph{IEEE Transactions on Control of Network Systems},
  2021.

\bibitem{littman1994markov}
M.~L. Littman, ``Markov games as a framework for multi-agent reinforcement
  learning,'' in \emph{Machine learning proceedings}.\hskip 1em plus 0.5em
  minus 0.4em\relax Elsevier, 1994, pp. 157--163.

\bibitem{kar2013cal}
S.~Kar, J.~M. Moura, and H.~V. Poor, ``${{\cal Q}{\cal D}} $-learning: A
  collaborative distributed strategy for multi-agent reinforcement learning
  through ${\rm consensus}+{\rm innovations} $,'' \emph{IEEE Transactions on
  Signal Processing}, vol.~61, no.~7, pp. 1848--1862, 2013.

\bibitem{macua2018diff}
S.~V. Macua, A.~Tukiainen, D.~G.-O. Hern{\'a}ndez, D.~Baldazo, E.~M. de~Cote,
  and S.~Zazo, ``Diff-dac: Distributed actor-critic for average multitask deep
  reinforcement learning,'' in \emph{Adaptive Learning Agents (ALA)
  Conference}, 2018.

\bibitem{sutton2011horde}
R.~S. Sutton, J.~Modayil, M.~Delp, T.~Degris, P.~M. Pilarski, A.~White, and
  D.~Precup, ``Horde: A scalable real-time architecture for learning knowledge
  from unsupervised sensorimotor interaction,'' in \emph{The 10th International
  Conference on Autonomous Agents and Multiagent Systems-Volume 2}, 2011, pp.
  761--768.

\bibitem{olshevsky2014linear}
A.~Olshevsky, ``Linear time average consensus on fixed graphs and implications
  for decentralized optimization and multi-agent control,'' \emph{arXiv
  preprint arXiv:1411.4186}, 2014.

\bibitem{Jing22ACC}
G.~Jing, H.~Bai, J.~George, A.~Chakrabortty, and P.~K. Sharma, ``Distributed
  cooperative multi-agent reinforcement learning with directed coordination
  graph,'' in \emph{2022 American Control Conference (ACC), to appear}.\hskip
  1em plus 0.5em minus 0.4em\relax IEEE, 2022.

\bibitem{qu2020scalablenips}
G.~Qu, Y.~Lin, A.~Wierman, and N.~Li, ``Scalable multi-agent reinforcement
  learning for networked systems with average reward,'' \emph{Advances in
  Neural Information Processing Systems}, vol.~33, 2020.

\bibitem{sunehag2017value}
P.~Sunehag, G.~Lever, A.~Gruslys, W.~M. Czarnecki, V.~Zambaldi, M.~Jaderberg,
  M.~Lanctot, N.~Sonnerat, J.~Z. Leibo, K.~Tuyls \emph{et~al.},
  ``Value-decomposition networks for cooperative multi-agent learning,''
  \emph{arXiv preprint arXiv:1706.05296}, 2017.

\bibitem{zhang2021fop}
T.~Zhang, Y.~Li, C.~Wang, G.~Xie, and Z.~Lu, ``Fop: Factorizing optimal joint
  policy of maximum-entropy multi-agent reinforcement learning,'' in
  \emph{International Conference on Machine Learning}.\hskip 1em plus 0.5em
  minus 0.4em\relax PMLR, 2021, pp. 12\,491--12\,500.

\bibitem{Fazel18}
M.~Fazel, R.~Ge, S.~Kakade, and M.~Mesbahi, ``Global convergence of policy
  gradient methods for the linear quadratic regulator,'' in \emph{International
  Conference on Machine Learning}.\hskip 1em plus 0.5em minus 0.4em\relax PMLR,
  2018, pp. 1467--1476.

\bibitem{malik2020derivative}
D.~Malik, A.~Pananjady, K.~Bhatia, K.~Khamaru, P.~L. Bartlett, and M.~J.
  Wainwright, ``Derivative-free methods for policy optimization: Guarantees for
  linear quadratic systems,'' \emph{Journal of Machine Learning Research},
  vol.~21, no.~21, pp. 1--51, 2020.

\bibitem{Hajine19}
D.~Hajinezhad, M.~Hong, and A.~Garcia, ``Zone: Zeroth-order nonconvex
  multiagent optimization over networks,'' \emph{IEEE Transactions on Automatic
  Control}, vol.~64, no.~10, pp. 3995--4010, 2019.

\bibitem{Gratton20}
C.~Gratton, N.~K. Venkategowda, R.~Arablouei, and S.~Werner,
  ``Privacy-preserving distributed zeroth-order optimization,'' \emph{arXiv
  preprint arXiv:2008.13468}, 2020.

\bibitem{Tang20}
Y.~Tang, J.~Zhang, and N.~Li, ``Distributed zero-order algorithms for nonconvex
  multi-agent optimization,'' \emph{IEEE Transactions on Control of Network
  Systems}, 2020.

\bibitem{Akhavan21}
A.~Akhavan, M.~Pontil, and A.~B. Tsybakov, ``Distributed zero-order
  optimization under adversarial noise,'' \emph{arXiv preprint
  arXiv:2102.01121}, 2021.

\bibitem{chen2021communication}
T.~Chen, K.~Zhang, G.~B. Giannakis, and T.~Basar, ``Communication-efficient
  policy gradient methods for distributed reinforcement learning,'' \emph{IEEE
  Transactions on Control of Network Systems}, 2021.

\bibitem{nesterov2017random}
Y.~Nesterov and V.~Spokoiny, ``Random gradient-free minimization of convex
  functions,'' \emph{Foundations of Computational Mathematics}, vol.~17, no.~2,
  pp. 527--566, 2017.

\bibitem{pirotta2015policy}
M.~Pirotta, M.~Restelli, and L.~Bascetta, ``Policy gradient in lipschitz markov
  decision processes,'' \emph{Machine Learning}, vol. 100, no.~2, pp. 255--283,
  2015.

\bibitem{kumar2020zeroth}
H.~Kumar, D.~S. Kalogerias, G.~J. Pappas, and A.~Ribeiro, ``Zeroth-order
  deterministic policy gradient,'' \emph{arXiv preprint arXiv:2006.07314},
  2020.

\bibitem{vemula2019contrasting}
A.~Vemula, W.~Sun, and J.~Bagnell, ``Contrasting exploration in parameter and
  action space: A zeroth-order optimization perspective,'' in \emph{The 22nd
  International Conference on Artificial Intelligence and Statistics}.\hskip
  1em plus 0.5em minus 0.4em\relax PMLR, 2019, pp. 2926--2935.

\bibitem{xiao2004fast}
L.~Xiao and S.~Boyd, ``Fast linear iterations for distributed averaging,''
  \emph{Systems \& Control Letters}, vol.~53, no.~1, pp. 65--78, 2004.

\bibitem{bhandari2019global}
J.~Bhandari and D.~Russo, ``Global optimality guarantees for policy gradient
  methods,'' \emph{arXiv preprint arXiv:1906.01786}, 2019.

\bibitem{agarwal2021theory}
A.~Agarwal, S.~M. Kakade, J.~D. Lee, and G.~Mahajan, ``On the theory of policy
  gradient methods: Optimality, approximation, and distribution shift.''
  \emph{Journal of Machine Learning Research}, vol.~22, no.~98, pp. 1--76,
  2021.

\bibitem{feng2019exponential}
H.~Feng and J.~Lavaei, ``On the exponential number of connected components for
  the feasible set of optimal decentralized control problems,'' in \emph{2019
  American Control Conference (ACC)}.\hskip 1em plus 0.5em minus 0.4em\relax
  IEEE, 2019, pp. 1430--1437.

\bibitem{zhang2021new}
Y.~Zhang, Y.~Zhou, K.~Ji, and M.~M. Zavlanos, ``A new one-point
  residual-feedback oracle for black-box learning and control,''
  \emph{Automatica}, p. 110006, 2021.

\bibitem{xiao2005scheme}
L.~Xiao, S.~Boyd, and S.~Lall, ``A scheme for robust distributed sensor fusion
  based on average consensus,'' in \emph{IPSN 2005. Fourth International
  Symposium on Information Processing in Sensor Networks, 2005.}\hskip 1em plus
  0.5em minus 0.4em\relax IEEE, 2005, pp. 63--70.

\end{thebibliography}
	\bibliographystyle{IEEEtran} % makes bibtex use IEEEtran style
	
	\begin{IEEEbiography}{Gangshan Jing} received the Ph.D. degree in Control Theory and Control Engineering from Xidian University, Xi'an, China, in 2018. From 2016-2017, he was a research assistant at Hong Kong Polytechnic University. From 2018 to 2019, he was a postdoctoral researcher at Ohio State University. From 2019 to 2021, he was a postdoctoral researcher  at North Carolina State University. Since 2021 Dec., he has been an assistant professor in School of Automation, at Chongqing University. His research interests include control, optimization, and machine learning for network systems.
	\end{IEEEbiography} 
	\vspace{-0.5cm}
	\begin{IEEEbiography}{He Bai} received his Ph.D. degree in Electrical Engineering from Rensselaer Polytechnic Institute, Troy, NY, in 2009. From 2009 to 2010, he was a postdoctoral researcher at Northwestern University, Evanston, IL. From 2010 to 2015, he was a Senior Research and Development Scientist at UtopiaCompression Corporation, Los Angeles, CA. In 2015, he joined the School of Mechanical and Aerospace Engineering at Oklahoma State University, Stillwater, OK, as an assistant professor. His research interests include distributed estimation, control and learning, reinforcement learning,  nonlinear control, and robotics. 
	\end{IEEEbiography} 
	\vspace{-0.5cm}
	\begin{IEEEbiography}{Jemin George} received his M.S. (’07), and Ph.D. (’10) in Aerospace Engineering from the State University of New York at Buffalo. Prior to joining ARL in 2010, he worked at the U.S. Air Force Research Laboratory’s Space Vehicles Directorate and the National Aeronautics, and Space Administration's Langley Aerospace Research Center. From 2014-2017, he was a Visiting Scholar at the Northwestern University, Evanston, IL. His principal research interests include decentralized/distributed learning, stochastic systems, control theory, nonlinear estimation/filtering, networked sensing and information fusion. 
	\end{IEEEbiography} 
	\vspace{-0.5cm}
	\begin{IEEEbiography}{Aranya Chakrabortty} received the Ph.D. degree in
		Electrical Engineering from Rensselaer Polytechnic Institute, NY in 2008. From 2008 to 2009 he was a postdoctoral research associate at University of
		Washington, Seattle, WA. From 2009 to 2010 he was an assistant professor at Texas Tech University, Lubbock, TX. Since 2010 he has joined the Electrical and Computer Engineering department at North Carolina State University, Raleigh, NC, where he is currently a Professor. His research interests are in all branches of control theory with applications to electric power systems. He received the NSF CAREER award in 2011.
	\end{IEEEbiography} 
	
	\begin{IEEEbiography}{Piyush Sharma} received his M.S. and Ph.D. degrees in Applied Mathematics from the University of Puerto Rico and Delaware State University respectively. He has government and industry work experiences. Currently, he is with the U.S. Army as an AI Coordinator at ATEC, earlier a Computer Scientist at DEVCOM ARL. Prior to joining ARL, he worked at Infosys’ Data Analytics Unit (DNA) as a Senior Associate Data Scientist responsible for Thought Leadership and solving stakeholders' problems.
	\end{IEEEbiography}

\end{document}